\pdfoutput=1
\documentclass[10pt]{article}

\usepackage[preprint]{paperstyle}

\usepackage{microtype}
\usepackage{graphicx}
\usepackage{subcaption}
\usepackage{wrapfig}
\usepackage{booktabs}
\usepackage{multirow}
\usepackage{url}
\usepackage{nicefrac}
\usepackage[table]{xcolor}

\usepackage{hyperref}

\usepackage{amsmath}
\usepackage{amssymb}
\usepackage{amsfonts}
\usepackage{mathtools}
\usepackage{amsthm}

\usepackage{enumitem}
\usepackage{thmtools}
\usepackage{thm-restate}

\usepackage{algorithm}
\usepackage{algorithmic}

\graphicspath{{figs/}}

\usepackage{amsmath,amsfonts,bm}

\def\rvb{{\boldsymbol{b}}}

\def\rvu{{\boldsymbol{i}}}

\def\rvu{{\boldsymbol{u}}}
\def\rvv{{\boldsymbol{v}}}
\def\rvw{{\boldsymbol{w}}}
\def\rvx{{\boldsymbol{x}}}
\def\rvy{{\boldsymbol{y}}}
\def\rvz{{\boldsymbol{z}}}

\def\normal{{\mathcal{N}}}

\newcommand{\EE}{\mathbb{E}}
\newcommand{\RR}{\mathbb{R}}

\newcommand{\KL}{D_{\mathrm{KL}}}
\newcommand{\FI}{D_{\mathrm{FI}}}

\def\mI{{\mathbf{I}}}

\def\mX{{\mathbf{X}}}

\def\mZ{{\mathbf{Z}}}

\newcommand{\df}{\mathrm{d}}
\newcommand{\norm}[1]{\left\lVert#1\right\rVert}

\definecolor{todopurple}{HTML}{9370DB}

\theoremstyle{plain}
\newtheorem{theorem}{Theorem}[section]

\newtheorem{lemma}[theorem]{Lemma}
\theoremstyle{definition}
\newtheorem{assumption}[theorem]{Assumption}
\theoremstyle{remark}
\newtheorem{remark}[theorem]{Remark}

\title{\texttt{Blade:} A Derivative-free Bayesian Inversion Method using Diffusion Priors}

\author{\name Hongkai Zheng\thanks{Equal contribution.} \email hzzheng@caltech.edu \\
      \addr California Institute of Technology
      \AND
      \name Austin Wang\footnotemark[1] \email akwang@caltech.edu \\
      \addr California Institute of Technology
      \AND
      \name Zihui Wu \email ray.wustl@gmail.com \\
      \addr California Institute of Technology
      \AND
      \name Daniel Zhengyu Huang \email huangdz@bicmr.pku.edu.cn \\
      \addr Peking University
      \AND
      \name Ricardo Baptista \email r.baptista@utoronto.ca \\
      \addr University of Toronto
      \AND
      \name Yisong Yue \email yyue@caltech.edu \\
      \addr California Institute of Technology}

\def\openreview{\url{https://openreview.net/forum?id=XXXX}}

\begin{document}

\maketitle

\makeatletter
\let\AND\@undefined
\makeatother

\begin{abstract}
Derivative-free Bayesian inversion arises in science and engineering applications, particularly when forward model is costly or infeasible to differentiate through.
Existing derivative-free methods collapse the posterior to a point estimate or return severely over-confident uncertainty on high-dimensional, nonlinear problems.
We introduce \texttt{Blade}, which produces accurate and well-calibrated posteriors using an ensemble of interacting particles.
\texttt{Blade} leverages diffusion models as data-driven priors, and only queries the forward model through forward evaluations (i.e., derivative-free).
Theoretically, we show the convergence and stability of \texttt{Blade} under forward model approximation and prior score estimation error.
Empirically, on nonlinear fluid dynamics, \texttt{Blade} produces well-calibrated posterior samples that existing derivative-free methods cannot, as measured by CRPS, the spread-skill ratio, and the rank histogram.
Its accuracy and calibration improve consistently with more iterations and particles, backed by our convergence and stability analysis and empirical experiments.
\end{abstract}

\section{Introduction}

Inverse problems, which seek to infer underlying system states or parameters from indirect and noisy observations, are fundamental to numerous applications in science and engineering. 
For instance, numerical weather prediction requires inferring the atmospheric state by assimilating observational data from weather stations and satellites~\citep{bannister2017review}.
Solving these problems requires overcoming three major challenges: first, they are often high-dimensional and ill-posed, meaning that the solution may be non-unique or unstable under perturbations~\citep{hadamard2014lectures}; second, the design of priors or regularizers is non-trivial and has a significant impact on the solution; third, the associated forward models may involve complicated numerical algorithms that make derivative calculation  impractical.
Indeed, the weather  example admits multiple possible solutions, demands carefully designed priors, and involves intricate numerical procedures with various \textbf{non-differentiable} steps like remapping, branching, and discrete search~\citep{park2013data,white2000ifs}. As such, derivative-free Bayesian inversion methods that use flexible priors to perform reliable uncertainty quantification for high dimensional problems are quite desirable~\citep{park2013data}.

We consider the inverse problems in the canonical form:
\begin{equation} \label{eq:inv problem}
    \rvy = \mathcal{G}(\rvx^*) + \boldsymbol{\epsilon},
\end{equation}
where  $\rvy \in \mathbb{R}^m$ is the observation or measurement, $\rvx^* \in \mathbb{R}^n$ is the unknown state, $\mathcal{G}$ is the forward model accessible only via forward evaluations (i.e., black-box access), and $\boldsymbol{\epsilon} \in \mathbb{R}^m$ is the measurement noise, often modeled as additive Gaussian $\mathcal{N}(0, \sigma_y^2 \mI)$. The Bayesian framework characterizes the solution as a posterior distribution $p(\rvx^*|\rvy) \propto p(\rvx^*)p(y|\rvx^*)$ that enables uncertainty quantification for principled decision-making~\citep{sanz2023inverse}.  As the gradient of $\mathcal{G}$ is difficult or impractical to compute, one typically resorts to derivative-free Bayesian inversion methods.


Traditional methods for derivative-free Bayesian inversion include Markov chain Monte Carlo (MCMC) methods~\citep{geyer1992practical, gelman1997weak, cotter2013mcmc} and Sequential Monte Carlo (SMC)~\citep{del2006sequential}. These methods offer convergence guarantees but face significant scalability challenges for high dimensional problems.  Approximate Bayesian methods~\citep{garbuno2020interacting, carrillo2022consensus, huang2022efficient} offer better efficiency, but often struggle to capture complex posteriors. Additionally, these methods require access to the prior density (up to a normalizing constant), which can be difficult to directly model in high dimensions.

Many recent derivative-free algorithms \citep{zheng2024ensemble, tang2024solving, huang2024symbolic} leverage diffusion models (DMs) as plug-and-play priors for solving high-dimensional inverse problems with complex prior distributions.  DMs can flexibly capture complex prior distributions from data, but require optimization or sampling for posterior inference, mainly due to modeling the score function rather than the density.  Optimization-based approaches disregard posterior spread and thus often fail to capture spread or uncertainty even in simple Gaussian settings with linear forward models (see Fig.~\ref{fig:linear-gmm}(a)). Sampling-based derivative-free approaches can be asymptotically correct~\citep{trippe2022diffusion,wu2023practical,cardoso2024monte, dou2024diffusion}, but have not yet demonstrated the capability to produce probabilistically calibrated samples in the high-dimensional nonlinear setting.

Our contributions are summarized as follows:
\vspace{-1pt}
\begin{itemize}[leftmargin=1.5em,itemsep=0.05em, topsep=0pt]
    \item We propose \texttt{Blade}, a derivative-free, ensemble-based Bayesian inversion algorithm that can produce well-calibrated posterior samples for inverse problems with diffusion prior. 
    \item We provide theoretical justification by establishing convergence and showing the stability of \texttt{Blade} under both statistical linearization and prior score estimation error with standard assumptions. 
    \item We evaluate \texttt{Blade} using various probabilistic verifications. In controlled settings we perform direct distributional checks against the ground truth posterior. In a challenging nonlinear fluid dynamics problem, we assess posterior quality using standard probabilistic metrics. Across all these tests, \texttt{Blade} demonstrates superior performance compared to competing approaches.
\end{itemize}

\section{Background}


\textbf{Diffusion Models.}
We consider diffusion models in the unified EDM framework \citep{karras2022elucidating}. Diffusion models define a forward stochastic process to evolve the original data distribution $p_0(\rvx)$ to an approximately Gaussian distribution $p_T(\rvx) = \mathcal{N}(0, s^2(T)\sigma(T)^2 \mI)$, where $\sigma(t)$ is a pre-defined noise schedule function and $s(t)$ is the pre-defined scaling function. Without loss of generality, we set $s(t)=1$ because every other schedule is equivalent to it up to a simple reparameterization as shown in~\citet{karras2022elucidating}. We consider the following form of denoising diffusion process: 
\begin{equation}
\label{eq:reverse-diffusion}
\df \rvx_t = -\left(2\dot{\sigma}(t) \sigma(t) + \beta(t)\right) \nabla_{\rvx_t} \log p\left(\rvx_t;\sigma(t)\right) \df t + \sqrt{2\dot{\sigma}(t) \sigma(t) + 2\beta(t)}\,\df \bar{\rvw}_t,
\end{equation}
where $\beta(t)$ can be any non-negative function as shown in~\citet{zhang2022fast}. 
Generating new samples from $p_0(\rvx)$ amounts to integrating Eq.~\eqref{eq:reverse-diffusion} from a random sample from $p_T(\rvx_T)$. This requires computation of the time-dependent score function $\nabla \log p(\rvx_t;\sigma(t))$, which can be approximated with a neural network: $
    s_{\theta}(\rvx_t, t) \approx \nabla \log p(\rvx_t;\sigma(t))$. In our work, we assume that we have access to such a pre-trained score function, which we will simply refer to as the diffusion model.


\textbf{Split Gibbs Sampling.}
The Split Gibbs Sampler (SGS) \citep{vono2019split} is a Markov chain Monte Carlo (MCMC) method that aims to sample the posterior  $p(\rvx \mid \rvy) \propto p(\rvy \mid \rvx) p(\rvx) = \exp(-f(\rvx; \rvy) - g(\rvx))$,
where $f(\rvx; \rvy) = -\log p(\rvy \mid \rvx)$ and $g(\rvx) = -\log p(\rvx)$. Instead of direct posterior sampling, SGS samples the auxiliary distribution:
\begin{equation}\label{eq:split dist}
\pi^{XZ}(\rvx, \rvz) \propto \exp\Bigl(-f(\rvz; \rvy) - g(\rvx) - \tfrac{1}{2 \rho^2} \norm{\rvx - \rvz}^2_2\Bigr).
\end{equation}
where $\rvz \in \mathbb{R}^n$ is an auxiliary variable and $\rho$ is a parameter that controls the distance between $\rvx$ and $\rvz$. As shown by \citet{vono2019split}, sampling the posterior becomes equivalent to sampling Eq.~\eqref{eq:split dist} as $\rho$ approaches $0$. Suppose $\rvx^{(0)}$ is the initial state and $k$ is iteration index. Sampling Eq.~\eqref{eq:split dist} is achieved using a Gibbs sampling procedure that alternates between the following two steps:
\begin{align}
\rvz^{(k)} &\sim \pi^{Z \mid X = \rvx^{(k)}}(\rvz) \propto \exp\Bigl(-f(\rvz; \rvy)- \tfrac{1}{2\rho^2} \norm{\rvx^{(k)} - \rvz}_2^2\Bigr), \label{eq:sgs_likelihood}\\
\rvx^{(k + 1)} &\sim \pi^{X \mid Z = \rvz^{(k)}}(\rvx) \propto \exp\Bigl(- g(\rvx) - \tfrac{1}{2\rho^2} \norm{\rvx - \rvz^{(k)}}_2^2\Bigr). \label{eq:sgs_prior}
\end{align}

\textbf{Diffusion-based Split Gibbs Sampling.}
Recent works have explored adapting generative model priors into the Gibbs sampling~\citep{janati2025mixture, achituveinverse} and split Gibbs framework~\citep{bouman2023generative, coeurdoux2023plug, xu2024provably, wu2024principled, chu2025split}. These methods focus on developing different ways to realize the prior step. This is orthogonal to our \texttt{Blade} method, which primarily focuses on the design of the likelihood step. For the prior step, we follow the practice of \citet{wu2024principled} which deduces the prior step to a denoising diffusion process within the EDM framework~\citep{karras2022elucidating}, enabling the use of any pre-trained diffusion model as the generative prior. For the likelihood step, the prior work either uses gradient-based Langevin Monte Carlo or derives the closed-form expression of $\pi^{Z|X}$ for linear problems, which requires the access to the gradient or adjoint operator of the forward model. In contrast, our \texttt{Blade} method implements the likelihood step as an interacting particle system which does not rely on derivative or adjoint operator information. 





\textbf{Ensemble Kalman Methods.}
Ensemble Kalman methodology was introduced by~\citet{evensen1994sequential} as a way of performing statistical linearization \citep{booton1954nonlinear}: estimating a surrogate linear model from samples (e.g., those generated using a black-box forward model).  This approach is appealing since it sidesteps the need to compute gradients from the generating process, and can then be integrated into various Bayesian inference schemes. In the inverse problem setting, this idea was used to develop optimization-based frameworks such the Ensemble Kalman Inversion (EKI) framework \citep{iglesias2013ensemble,kovachki2018eki,iglesias2016regularizing, chada2020tikhonov,huang2022iterated} as well as derivative-free diffusion guidance methods \citep{kim2024derivative, zheng2024ensemble}. \texttt{Blade} builds upon the  fundamental idea of statistical linearization of a black-box forward model, and carefully incorporates it into the split Gibbs  framework with a diffusion prior to produce well-calibrated posterior samples for high-dimensional problems.

\section{Method}\label{sec:method}

\begin{figure}[t]
    \centering
    \includegraphics[width=0.77\textwidth]{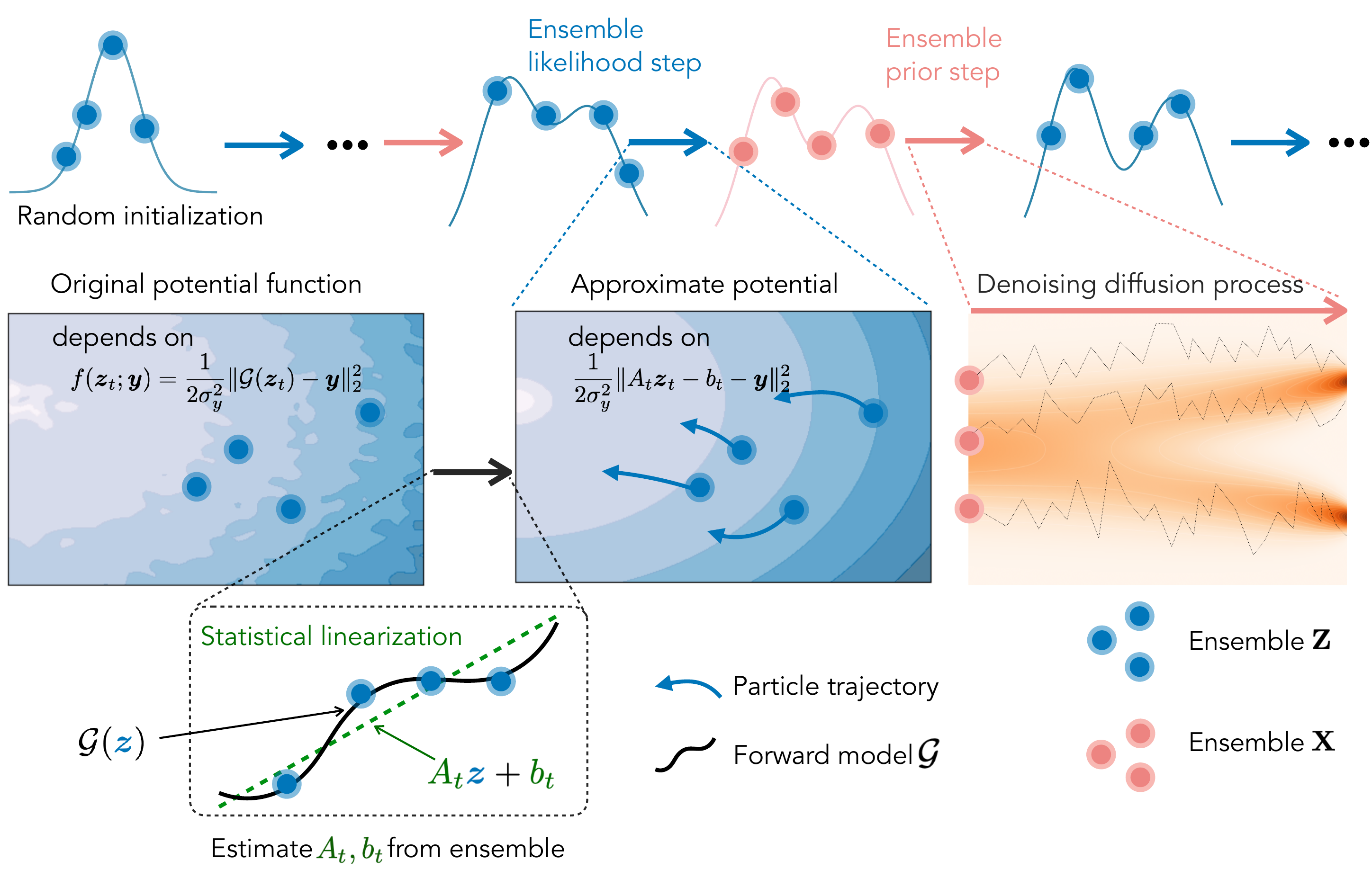}
    \caption{Illustrative depiction of \texttt{Blade} (see Sec.~\ref{sec:method}).}
    \label{fig:diagram}
\end{figure}

\begin{figure}[t]
\centering
\begin{minipage}[t]{0.6\textwidth}
\vspace{-5pt}
\begin{algorithm}[H]
\caption{\texttt{Blade} algorithm}
\label{alg:Blade}
\begin{algorithmic}
\STATE {\bfseries Input:} initial ensemble $\mathbf{X}^{(0)} = \{\mathbf{x}^{(j)} \in \mathbb{R}^n\}_{j=1}^{J}$,
\STATE \hspace{1em}number of iterations $K$, schedule $\{\rho_k\}_{k=0}^{K-1}$,
\STATE \hspace{1em}observed data $\mathbf{y} \in \mathbb{R}^m$, pre-trained diffusion model $s_{\theta}$
\FOR{$k \in \{0,\dots, K - 1\}$}
\STATE $\mathbf{Z}^{(k)} \gets \operatorname{EnsembleLikelihoodStep}(\mathbf{X}^{(k)}, \rho_k)$ \\ \COMMENT{See Alg.~\ref{alg:likelihood-step}}
\STATE $\mathbf{X}^{(k+1)} \gets \operatorname{Ensemble-prior-step}(\mathbf{Z}^{(k)}, s_{\theta}, \rho_k)$ \\ \COMMENT{See Alg.~\ref{alg:prior-step}}
\ENDFOR
\STATE \textbf{Return} $\mathbf{X}^{(K)}$
\end{algorithmic}
\end{algorithm}
\end{minipage}
\hspace{1pt}
\begin{minipage}[t]{0.3\textwidth}
\vspace{0pt}
\centering
\includegraphics[width=\linewidth]{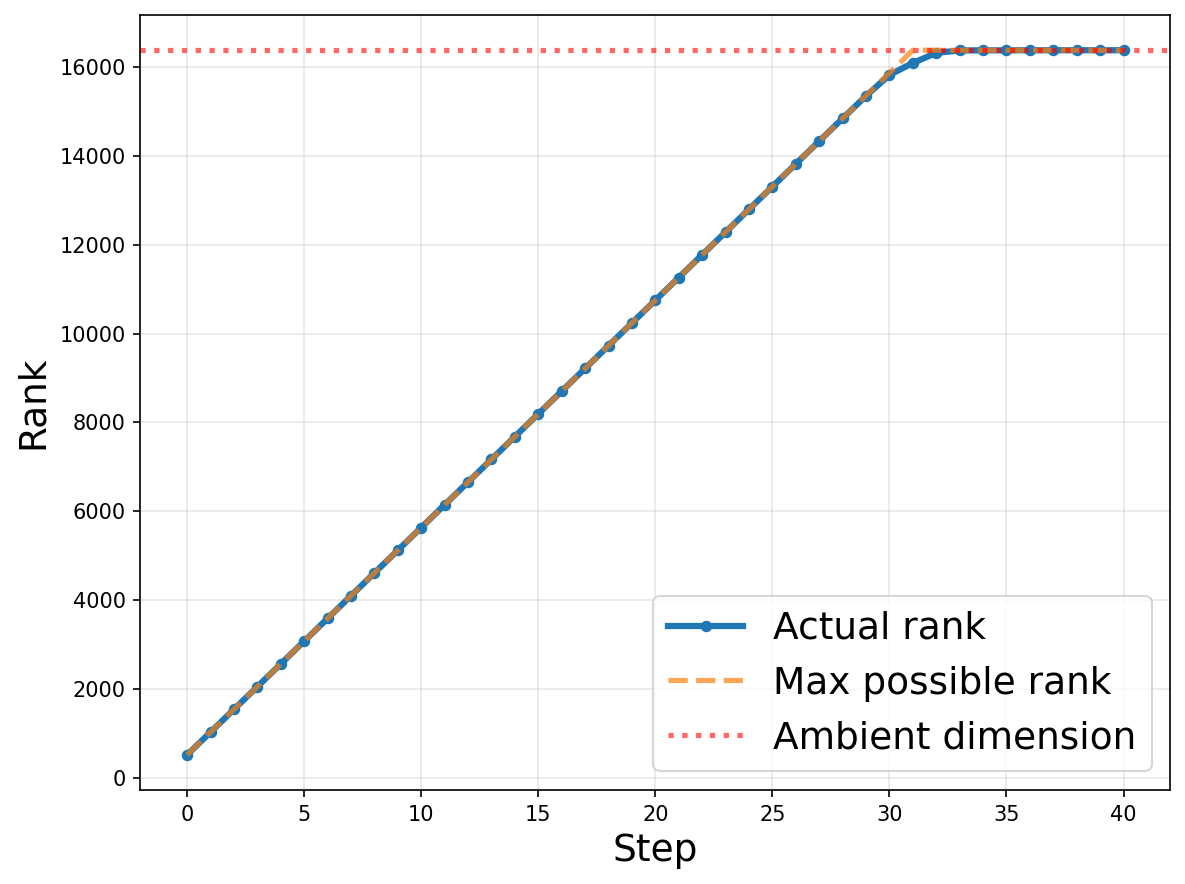}
\captionof{figure}{Evolution of the rank of the space spanned by ensemble particles during \texttt{Blade} iterations.}
\label{fig:ensemble_rank}
\end{minipage}
\end{figure}

 
The name \texttt{Blade} is derived from the key components of the algorithm: Bayesian inversion, Linearization, Alternating updates, Derivative-free, and Ensembling. As illustrated in Fig.~\ref{fig:diagram} and summarized in Alg.~\ref{alg:Blade},  \texttt{Blade} is built on the split Gibbs framework and evolves a set of interacting particles that alternate between a derivative-free likelihood sampling step and a denoising diffusion prior step. In the likelihood step, the original (and potentially complex) potential is approximated by a smooth quadratic form through statistical linearization, which enables derivative-free sampling (Sec.~\ref{sec:likelihood}).
In the prior step, the ensemble member operates independently following denoising diffusion process (Sec.~\ref{sec:prior}).
Sec.~\ref{sec:practical-algo} presents the full algorithm along with practical considerations. Finally, Sec.~\ref{sec:theory} provides theoretical justification by establishing convergence and stability under different modeling error.

\subsection{Derivative-free likelihood step via statistical linearization}\label{sec:likelihood}
Let  $\mX^{(k)}=\{\rvx^{(j)}\}_{j=1}^J$ denote the ensemble of $J$ particles at $k$-th alternating iteration of the SGS framework. In the likelihood step (Eq.~\eqref{eq:sgs_likelihood}), we aim to sample $\rvz^{(j)}$ from $\pi^{Z|X=\rvx^{(j)}}(\rvz) \propto \exp(-f(\rvz;\rvy) - \frac{1}{2\rho^2}\|\rvz - \rvx^{(j)}\|_2^2)$ for each $j\in \{1,\dots, J\}$.
Our starting point is the covariance-preconditioned Langevin dynamics with the large particle limit, which is known to have improved conditioning and convergence under quadratic potentials~\citep{reich2021fokker,garbuno2020interacting}:
\begin{equation}\label{eq:cov-precond-langevin}
\df \rvz_t^{(j)} = - C_t\nabla \left(f(\rvz^{(j)}_t;\rvy)+ \frac{1}{2\rho^2}\|\rvz^{(j)}_t - \rvx^{(j)}\|_2^2\right)\df t + \sqrt{2C_t}\,\df \rvw_t,
\end{equation}
where $q_t$ is the particle distribution,  
$\bar{\rvz}_t:=\EE_{q_t} [\rvz_t]$, and $C_t:=\EE_{q_t} [(\rvz_t-\bar{\rvz}_t)(\rvz_t-\bar{\rvz}_t)^\top]$.
As shown in Lemma~\ref{lemma1}, Eq.~\eqref{eq:cov-precond-langevin} admits $\pi^{Z|X=\rvx^{(j)}}(\rvz)$ as its stationary distribution, under the mild assumption that the particle distribution does not collapse to a Dirac measure.  For the inverse problem in Eq.~\eqref{eq:inv problem}, we have $
f(\rvz^{(j)}_t;\rvy) = \frac{1}{2\sigma_y^2}\|\mathcal{G}(\rvz^{(j)}_t)-\rvy\|_2^2 $. Therefore, running Eq.~\eqref{eq:cov-precond-langevin} relies on the derivative of the forward model $\mathcal{G}$, which may not be available. To circumvent this, we approximate $\mathcal{G}$ with a linear surrogate model $\rvy =A_t\rvz_t + \rvb_t $ with the minimal least square error defined by: 
\begin{equation}
     \min_{A_t,\rvb_t} \EE_{q_t} \|\mathcal{G}(\rvz_t) - (A_t\rvz_t + \rvb_t)\|_2^2.
\end{equation}
Setting the derivatives w.r.t. $A_t$ and $\rvb_t$ to zero gives the closed-form solution: 
\begin{equation}\label{eq:linearization-solution}
\begin{aligned}
A_t
&= \EE_{q_t}\Bigl[\bigl(\mathcal{G}(\rvz_t) - \EE_{q_t}\mathcal{G}(\rvz_t)\bigr)(\rvz_t-\bar{\rvz}_t)^\top\Bigr] C_t^{-1},\\
\rvb_t
&= \EE_{q_t}\mathcal{G}(\rvz_t) - \EE_{q_t}A_t\rvz_t,
\end{aligned}
\end{equation}
where $C^{-1}_t$ is the pseudo-inverse of the covariance matrix.
This approach, known as statistical linearization, was first introduced in~\citet{booton1954nonlinear} and recently used in~\citet{kim2024derivative,zheng2024ensemble}.  Statistical linearization is exact when $\mathcal{G}$ is linear. Let $D\mathcal{G}$ denote the Jacobian of $\mathcal{G}$ and $\Tilde{\rvz}_t= \rvz_t - \bar{\rvz}_t$. Replacing $D\mathcal{G}$ with $DA_t=A_t$ in $     \nabla f(\rvz_t^{(j)};\rvy)$ yields:
\begin{equation}\label{eq:f-approx}
\nabla f(\rvz_t^{(j)};\rvy)
= \frac{1}{\sigma_y^2}(D^\top\mathcal{G})\bigl(\mathcal{G}(\rvz_t^{(j)})-\rvy\bigr)
\approx \frac{1}{\sigma_y^2}C_t^{-1}
\EE_{q_t}\Bigl[\Tilde{\rvz}_t\bigl(\mathcal{G}(\rvz_t)-\EE_{q_t}\mathcal{G}(\rvz_t)\bigr)^\top\Bigr]
\bigl(\mathcal{G}(\rvz_t^{(j)})-\rvy\bigr).
\end{equation}
Substituting Eq.~\eqref{eq:f-approx} into Eq.~\eqref{eq:cov-precond-langevin} gives us  the derivation of the likelihood step: 
\begin{equation}\label{eq:approx-dynamic}
\df \rvz_t^{(j)} = - \Bigl[\frac{1}{\sigma_y^2} \EE_{q_t}\bigl[\Tilde{\rvz}_t\bigl(\mathcal{G}(\rvz_t)-\EE_{q_t}\mathcal{G}(\rvz_t)\bigr)^\top\bigr] \bigl(\mathcal{G}(\rvz_t^{(j)})-\rvy\bigr) + \frac{1}{\rho^2} C_t\bigl(\rvz_t^{(j)}-\rvx^{(j)}\bigr)\Bigr] \df t + \sqrt{2C_t}\,\df \rvw_t,
\end{equation}

which does not require explicit computation of $\hat{A}$ and thus avoids computing the matrix pseudo-inverse. Eq.~\eqref{eq:approx-dynamic} also eliminates the need for derivatives of the forward model, allowing us to run the algorithm with only black-box access to $\mathcal{G}$. The dynamics in Eq.~\eqref{eq:approx-dynamic} shares a similar structure as that of Ensemble Kalman Sampling (EKS)~\citep{garbuno2020interacting} and ALDI~\citep{garbuno2020affine}. 
However, a key distinction is that each particle in \texttt{Blade} has its own target distribution associated with $\rvx^{(j)}$---an individualized, locally-adapted update rather than the single shared potential used by EKS and ALDI---letting the ensemble cover distinct posterior regions. 


\subsection{Ensemble-based prior step via denoising diffusion}\label{sec:prior}
Let  $\mZ^{(k)}=\{\rvz^{(j)}\}_{j=1}^J$ denote the ensemble of $J$ particles at $k$-th alternating iteration of the SGS framework. In the prior step (Eq.~\eqref{eq:sgs_prior}), we aim to sample $\rvx^{(j)}$ from $\pi^{X|Z=\rvz^{(j)}}(\rvx) \propto \exp(-g(\rvx) - \frac{1}{2\rho^2}\|\rvx - \rvz^{(j)}\|_2^2)$ for each $j\in \{1,\dots, J\}$.
As shown in~\citet{coeurdoux2023plug, wu2024principled}, the prior step can be formulated as denoising diffusion process. Specifically, recall that the forward process gives $p(\rvx_t \mid \rvx_0) = \mathcal{N}(\rvx_t; \rvx_0, \sigma(t)^2 \mI)$ under the EDM \citep{karras2022elucidating} framework, where $s(t) = 1$. By Bayes' theorem, we have:
\begin{equation}\label{prior_posterior}
p(\rvx_0 \mid \rvx_t) \propto p(\rvx_t \mid \rvx_0)\, p(\rvx_0) \propto \exp\Bigl(-g(\rvx_0) - \tfrac{1}{2 \sigma^2(t)} \norm{\rvx_0 - \rvx_t}^2_2\Bigr).
\end{equation}
By comparing Eq.~\eqref{prior_posterior} with the target distribution $\pi^{X|Z=\rvz^{(j)}}(\rvx)$, we can see that if $\rho = \sigma(t)$ and $\rvx_t = \rvz^{(j)}$, sampling from $\pi^{X|Z=\rvz^{(j)}}(\rvx)$ is equivalent to sampling from $p(\rvx_0|\rvx_t=\rvz^{(j)})$ . Therefore, the prior step of $j$-th particle can be implemented as the standard reverse process of the diffusion model given by Eq.~\eqref{eq:reverse-diffusion} starting from $\rvz^{(j)}$ at time $t^*$ where $t^*$ is chosen so that $\sigma(t^*)=\rho$.

\subsection{Practical algorithm}\label{sec:practical-algo}

\textbf{Practical implementation of likelihood step.} To implement the dynamics in Eq.~\eqref{eq:approx-dynamic} with a finite-particle system, we introduce two practical variants: \texttt{Blade(main)} and \texttt{Blade(diag)}. The former is our main algorithm, designed to preserve correct posterior uncertainty, while the latter often provides sharper point estimates. The pseudocode is provided in Algorithm~\ref{alg:likelihood-step} in the appendix.

\texttt{Blade (main)} ensures that the invariant measure of the finite-particle system remains the same as that of Eq.~\eqref{eq:approx-dynamic}. As shown in~\citet{nusken2019note}, the covariance-preconditioned stochastic process requires additional correction term as the diffusion term depends on the evolving particle. For $j$-th particle, we add a correction term to the drift of Eq.~\eqref{eq:approx-dynamic}, yielding:
\begin{equation}\label{eq:likelihood-dynamics-correction}
\df \rvz_t^{(j)} = - \Bigl[\sigma_{\rvy}^{-2} C_t A_t^\top\bigl(\mathcal{G}(\rvz_t^{(j)})-\rvy\bigr) + \rho^{-2} C_t\bigl(\rvz_t^{(j)}-\rvx^{(j)}\bigr)\Bigr] \df t + \sqrt{2C_t}\,\df \rvw_t + \tfrac{n+1}{J}\bigl(\rvz_t^{(j)} - \bar{\rvz}_t\bigr) \df t,
\end{equation}
where $n$ is the dimensionality of $\rvz$, $J$ is the ensemble size, and $\rvw_t \in \RR^J$. Lemma~\ref{lemma:finite-correction} verifies that Eq.~\eqref{eq:likelihood-dynamics-correction} has an invariant measure that is identical to that of Eq.~\eqref{eq:approx-dynamic}. Intuitively, the correction term $\frac{n+1}{J}(\rvz_t^{(j)} - \bar{\rvz}_t)$ pushes the particles away from each other and vanishes when $J \gg n$. For the computation of $\sqrt{C_t}$, we use the construction proposed in~\citet{garbuno2020affine} where $\sqrt{C_t} = \frac{1}{\sqrt{J}}(\rvz_t^{(1)} - \bar{\rvz}_t, \dots, \rvz_t^{(J)} - \bar{\rvz}_t) \in \RR^{n\times J}$, which avoids explicit matrix square roots.

\texttt{Blade (diag)} simply approximates the $\sqrt{C_t}$ by the diagonal standard deviation. This simplification introduces a force that pulls the particles into each other, resulting in a smaller spread. It often yields sharper point estimates which can lead to over-confidence. 


\textbf{Implementation of prior step.} Detailed pseudocode for the prior step can be found in Algorithm~\ref{alg:prior-step}. Note that in Algorithm~\ref{alg:prior-step}, $\mX_i$ represents the ensemble of particles at $t_i$ step, and the updates for all particles can be computed in parallel. For discretization, we use the Euler method with the step size scheme in~\citet{karras2022elucidating}. Further implementation details are deferred to Appendix~\ref{sec:resample}.


\textbf{Putting it all together.}
We provide pseudocode for the complete sampling algorithm in Algorithm~\ref{alg:Blade}. The method operates by iteratively updating an ensemble of particles, alternating between the likelihood and prior steps discussed above. At the same time, the parameter $\rho$ follows an annealing schedule that gradually decreases towards zero. Annealing $\rho$ from large to small implements a smooth path from a smooth posterior with good mixing to the sharp one with improved accuracy. Further details are deferred to Appendix~\ref{sec:implementation-details}.

\begin{remark}
    Like all other ensemble Kalman methods, when the ensemble size is finite, the statistical linearization in a single likelihood step of \texttt{Blade} is confined to a subspace spanned by the particles, meaning that it is typically not a full-rank update, which is sometimes problematic in other ensemble Kalman frameworks.  However, since \texttt{Blade} alternates between likelihood and prior steps, the randomness and nonlinearity of the prior step generally implies that sampling trajectory will explore all dimensions (assuming a full-rank prior).   
    To verify this empirically, we track the dimensionality of explored space over iterations in the setting where the ensemble size (512) is significantly smaller than the ambient dimension (16384). At iteration $k$, we form the matrix whose columns are the concatenated particles from all iterations up to $k$: $[X^{0}, Z^{0}, \dots, X^{(k)}, Z^{(k)}]$. The accumulated rank is the rank of this matrix.
    Fig.~\ref{fig:ensemble_rank} shows that
    the actual rank closely tracks the max possible rank and it reaches the ambient dimension within 35 steps, indicating that \texttt{Blade} has sufficient posterior exploration without the low-dimensional confinement.
\end{remark}

\subsection{Theoretical analysis} \label{sec:theory}

To provide theoretical grounding for Blade, we analyze the behavior of \texttt{Blade} through the lens of its continuous-time and large particle limit for the ease of understanding. In practice, \texttt{Blade} incurs two bias terms: $\epsilon_{\mathrm{model}}$ from the statistical linearization and $\epsilon_{\mathrm{score}}$ from the learned diffusion prior. By extending  existing proof techniques from~\citet{wu2024principled,vempala2019rapid,sun2024provable} to the interacting particle system with state-dependent preconditioner, our analysis quantifies how these errors together with the number of iterations $K$ affects the deviation from the reference process. Technical definitions and notations are collected in Appendix~\ref{sec:notation}.

\begin{restatable}[Stationary distribution]{theorem}{theoremStationary} \label{thm:stationary}
    Given any $\rho>0$, consider the oracle split-Gibbs algorithm that alternates between the likelihood step defined in Eq.~\eqref{eq:cov-precond-langevin} and the prior step defined in Eq.~\eqref{eq:reverse-diffusion} where each step is implemented perfectly without approximations. If the particle distribution is not a Dirac measure, then $\pi^{XZ}$ is a stationary distribution. 
    Furthermore, if the preconditioner $C_t$ is positive definite, $\pi^{XZ}$ is the unique stationary distribution.  
\end{restatable}

We defer the proof to Appendix~\ref{sec:proofs}.
\begin{remark}\label{sec:thm1-remark}
    Theorem~\ref{thm:stationary} shows that, when every sub-step is exact, $\pi^{XZ}$ is the stationary distribution under the standard argument with positive definite preconditioner. Note that Theorem~\ref{thm:stationary} holds for any positive definite preconditioner and the covariance matrix is just one special case. The preconditioner does not change the stationary distribution but rather specifies the geometry of the dynamics (See Kalman-Wasserstein gradient flow in~\citet{garbuno2020interacting}). This observation will be used in Theorem~\ref{thm:convergence} to compare two processes within a common metric structure. 
\end{remark}

The next theorem establishes convergence and stability of \texttt{Blade} under two practical approximation errors: $\epsilon_{\mathrm{model}}$ from the statistical linearization defined in Assumption~\ref{assumption:linearization} and $\epsilon_{\mathrm{score}}$ from the learned diffusion prior defined in Assumption~\ref{assumption:score}, under standard assumptions listed in Appendix~\ref{sec:proofs}. 




\begin{restatable}[Convergence analysis]{theorem}{theoremConvergence}
\label{thm:convergence}
    Given $\rho >0$, consider the following two processes that alternate between the likelihood step with horizon $t^\dagger$ and the prior step with horizon $t^*$, where $\sigma(t^*) = \rho$: 
    \begin{itemize}[leftmargin=1.5em]
        \item The approximate process that implements the likelihood step as in Eq.~\eqref{eq:approx-dynamic} (with forward model approximation) and the prior step as in Eq.~\eqref{eq:reverse-diffusion} (with diffusion model score approximation). Let $\Tilde{\mu}_t$ denote its distribution at time $t$, $C_t$ the associated covariance matrix, $\lambda^*_t$ the smallest non-zero eigenvalue of $C_t$. 
        \item The reference process that starts from the stationary distribution $\pi^{XZ}$, implements the likelihood step as Eq.~\eqref{eq:cov-precond-langevin} with the preconditioner $C_t$, and the prior step which runs Eq.~\eqref{eq:reverse-diffusion},  assuming exact knowledge of both the prior score function and forward model derivative. Let $\mu_t$ denote its distribution at time $t$. 
    \end{itemize}
    Let $T_k=k(t^\dagger +t^*), k=0,\dots, K$, $\lambda^*=\inf_{t\in\cup_k [T_k,T_k+t^\dagger]} \lambda^*_t$, and $\delta = \inf_{t\in[0,t^*]} \delta(t)$ where $\delta(t)$ is the diffusion term defined in Eq.~\eqref{eq:diffusion}. We assume both $\lambda^*$ and $\delta$ are strictly positive. We denote by $\epsilon_{\mathrm{score}}$ the score approximation error of the diffusion model defined in Assumption~\ref{assumption:score}, and $\epsilon_{\mathrm{model}}$ the forward model derivative approximation error defined in Assumption~\ref{assumption:linearization}.
    Assuming that $\KL(\pi^X||\mu_0)<+\infty$ and Assumption~\ref{assumption-null-space-invariance} holds,  for $K$ split Gibbs iterations, we have
    \begin{align}\label{eq:convergence-thm}
        \frac{1}{T_K} \int_0^{T_K} \FI(\mu_t\|\Tilde{\mu}_t)\,\df t 
        \leq \frac{4}{\min(\lambda^*,\delta)}
    \Biggl[
    \frac{\KL(\pi^X\|\Tilde{\mu}_0)}{K(t^\dagger + t^*)}
    + \frac{t^\dagger\epsilon_{\mathrm{model}} + t^*\epsilon_{\mathrm{score}}}{t^\dagger + t^*}
    \Biggr].
    \end{align}
    where $\FI$ and $\KL$ are Fisher divergence and KL divergence respectively, defined in Appendix~\ref{sec:notation}. 
\end{restatable}
The proof of is deferred to Appendix~\ref{sec:proofs}.

\begin{remark}\label{remark:theorem-extend}
Theorem~\ref{thm:convergence} provides two main insights. The first is the convergence to the reference process. The time-average Fisher divergence between the approximate process and reference process decays at an $O(1/K)$ rate up to a weighted sum of approximation errors. Second, the algorithm remains stable under the statistical linearization error and the prior score approximation error and these terms do not vanish with more iterations as expected.

\textbf{Interpretation of the reference process.}
The reference process stays at the target stationary distribution if the likelihood step is exact regardless of the chosen preconditioner $C_t$. The role of preconditioner $C_t$ here is to specify the geometry with respect to which we measure approximation error as mentioned in Remark~\ref{sec:thm1-remark}. See Kalman-Wasserstein gradient flow in~\citet{garbuno2020interacting} for how geometry depends on the preconditioner. Using $C_t$ from the approximate process to construct the reference process does not alter the target stationary distribution; it simply allows us to compare the two dynamics within a common metric structure. Hence, the convergence to the reference process indicates the convergence to the desired stationary distribution.

\textbf{Comparison to prior work.}
While the bound in Eq.~\eqref{eq:convergence-thm} is structurally similar to that of \texttt{PnPDM}~\citep{wu2024principled}, our analysis strictly generalizes the prior results to a more realistic and technically challenging regime. First, our method considers a different and more complex likelihood step dynamics with interacting particles and state-dependent preconditioner, whereas \citet{wu2024principled} considers the standard Langevin dynamics. Second, our setting is more realistic by considering finite time execution of the dynamics as well as the forward model error. 
Third, Theorem~\ref{thm:convergence} shows how the algorithm remains stable under forward model error and how the finite time horizon affect convergence, which are not addressed in the prior work.
\end{remark}


\begin{figure}[t]
    \centering
    \includegraphics[width=0.97\textwidth]{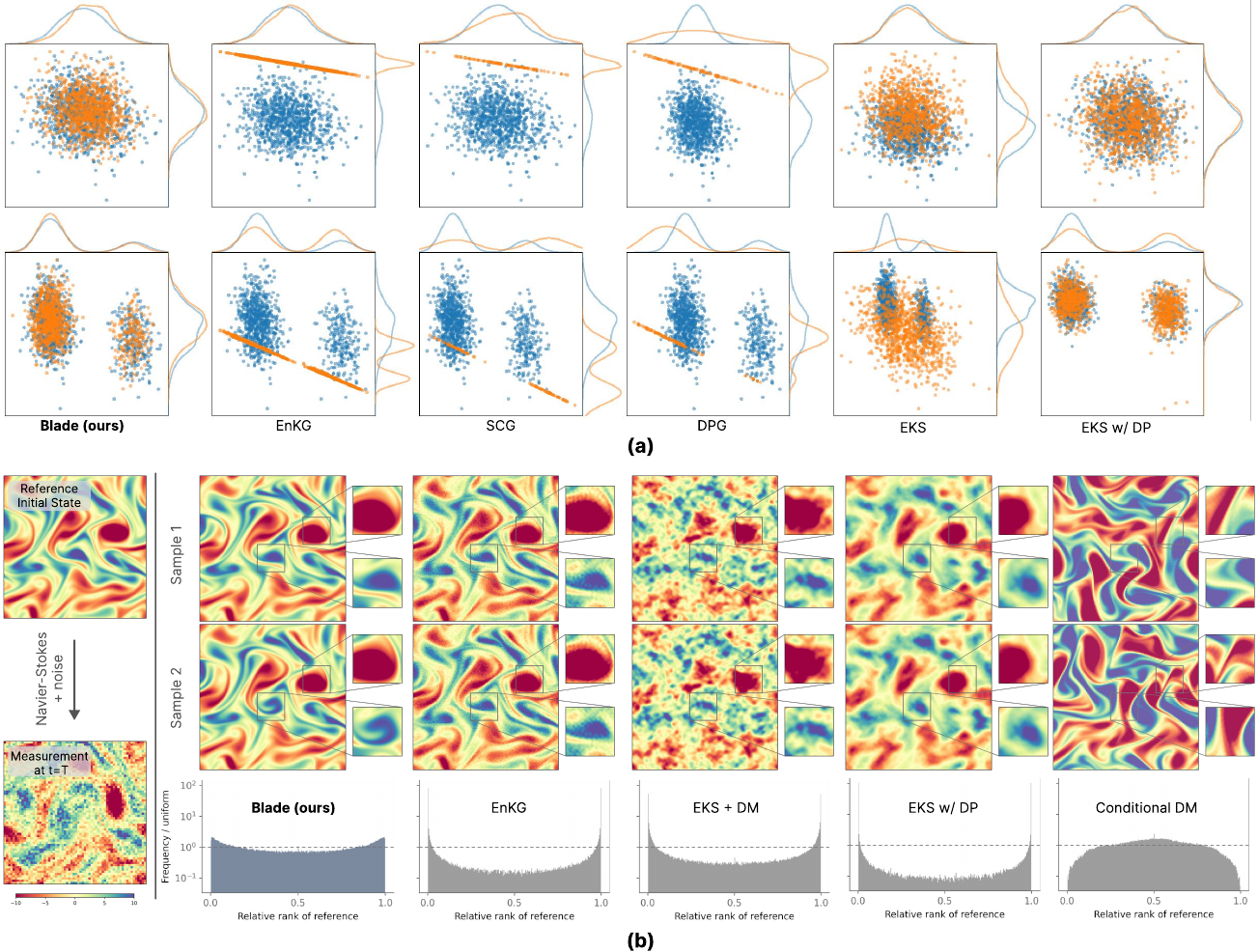}
    \caption{
        \texttt{Blade} produces multimodal and well-calibrated posterior samples across controlled and high-dimensional inverse problems.
  \textbf{(a)} Controlled two-dimensional inverse problems with analytic posteriors: a linear Gaussian posterior (top row) and a two-component non-isotropic Gaussian mixture posterior (bottom row). 
  Blue points are ground-truth posterior samples; orange points are samples from each method. \texttt{Blade} best matches the analytic posterior in both settings.
  \textbf{(b)} Navier--Stokes inverse problem. The left column shows the reference initial vorticity, the forward observation process, and the noisy later-time measurement. 
  Each method column shows two posterior draws (with zoom-ins) and a rank histogram computed over test cases and grid points. The histogram reports the relative rank of the reference value among the posterior ensemble, normalized by a uniform histogram. Flat (dashed line) indicates ideal calibration, U-shaped is under-dispersed, and inverted-U is over-dispersed. 
  \texttt{Blade} shows realistic local variability and the closest-to-uniform histogram among the shown methods. DP: diffusion prior; DM: diffusion model.
  }
    \label{fig:linear-gmm}
\end{figure}



\section{Experiments}

Inverse problems are ill-posed, so an ensemble of posterior samples with calibrated uncertainty is often more desirable than a single point estimate. We therefore evaluate \texttt{Blade} through probabilistic verification of its posterior samples. Sec.~\ref{sec:linear-gaussian} considers fully controlled settings with analytic ground-truth posteriors, enabling direct distributional checks. Sec.~\ref{sec:ns-experiment} turns to a challenging high-dimensional problem based on the Navier-Stokes equation where the ground-truth posterior is unknown but probabilistic verification methods are available. Sec.~\ref{sec:ablation-study} studies the effect of \texttt{Blade}'s hyperparameters and its robustness to diffusion prior choice. Sec.~\ref{sec:exp-on-point-estimate} reports results on popular point-estimate benchmarks as complementary evidence of \texttt{Blade}’s breadth.

\begin{table}[t]
    \centering
    \caption{Comparison on the Navier-Stokes inverse problem. The primary probabilistic metrics are CRPS and SSR. Rel L2 error (relative L2 error) is deterministic, included as a complementary metric.  $-$ indicates either that probabilistic metrics are inapplicable (deterministic method) or that it is too costly to generate enough samples from the algorithm for reliable calculation of probabilistic metrics. CDM-CA: conditional diffusion model with cross attention. CDM-Cat: conditional diffusion model with channel concatenation.}
    \resizebox{0.99\linewidth}{!}{
    {
    \begin{tabular}{l*{3}{ccc}}\toprule
         & \multicolumn{3}{c}{$\sigma_{\mathrm{noise}}=0$} & \multicolumn{3}{c}{$\sigma_{\mathrm{noise}}=1.0$} & \multicolumn{3}{c}{$\sigma_{\mathrm{noise}}=2.0$} \\ 
\cmidrule(lr){2-4} \cmidrule(lr){5-7} \cmidrule(lr){8-10}
         & CRPS$\downarrow$ & SSR$\rightarrow 1$ & Rel L2 error$\downarrow$ & CRPS$\downarrow$ & SSR$\rightarrow 1$ & Rel L2 error$\downarrow$ & CRPS$\downarrow$ & SSR$\rightarrow 1$ & Rel L2 error$\downarrow$ \\ \midrule 
        \textbf{Paired data} \\
CDM-CA & 2.900 & 0.983 & 1.362 & 2.872 & 1.059 & 1.409 & 2.993 & 1.087 & 1.542 \\
CDM-Cat & 1.413 & 0.896 & 0.653 & 1.805 & 0.979 & 0.873 & 2.211 & 0.974 & 1.043 \\
U-Net & --- & --- & 0.585 & --- & --- & 0.702 & --- & --- & 0.709 \\ 
\midrule 
\textbf{Unpaired data}\\
EKI     & 2.303 & 0.012 & 0.577 & 2.350 & 0.118 & 0.586 & 2.700 & 0.011 & 0.673 \\ 
EKS + DM   & 1.900 & 0.181 & 0.539 & 2.088 & 0.218 & 0.606 & 2.255 & 0.280 & 0.685 \\
EKS w/ DP & 1.280 & 0.061 & 0.336 &  1.723 & 0.085 & 0.455 & 2.094 & 0.080  & 0.547  \\
Localized EKS w/ DP & 1.643 & 0.056 & 0.428 & 1.887 & 0.081 & 0.495 & 2.057 & 0.089 & 0.542 \\
FKD & 1.604 & 0.002 & 0.399 & 1.416 & 0.050 & 0.368  & 1.810 & 0.012& 0.455 \\
DPG     & --- & --- & 0.325 & --- & --- & 0.408 & --- & --- & 0.466 \\
SCG     & --- & --- & 0.961 & --- & --- & 0.928 & --- & --- & 0.966 \\
EnKG    & 0.395 & 0.164 & 0.120 & 0.651 & 0.154 & 0.191 & 1.032 & 0.144 & 0.294 \\ 
\rowcolor{gray!20} \texttt{Blade (diag)} & 0.276 & 0.086 & \textbf{0.080} & 0.542 & 0.177 & \textbf{0.162} & 0.758 & 0.129 & \textbf{0.217} \\
\rowcolor{gray!20} \texttt{Blade (main)} & \textbf{0.216} & \textbf{0.955} & 0.110  & \textbf{0.453} & \textbf{0.950} & 0.229 & \textbf{0.608} & \textbf{0.949} & 0.306 \\
\bottomrule
\end{tabular}}}
\label{tab:navier_stokes}
\end{table}

\subsection{Gaussian and Gaussian mixture}\label{sec:linear-gaussian}
To enable direct distributional check, we consider two cases where the exact posteriors can be derived: (i) a linear Gaussian posterior and (ii) a two‐component non-isotropic Gaussian mixture posterior. For both we derive the analytic posterior and draw “ground‐truth” samples, then compare to those produced by different algorithms. We compare \texttt{Blade} against the existing derivative-free methods including \texttt{DPG}~\citep{tang2024solving}, \texttt{SCG}~\citep{huang2024symbolic}, \texttt{EnKG}~\citep{zheng2024ensemble}, EKS~\citep{garbuno2020interacting}, and EKS w/ DP (EKS with diffusion prior). Fig.~\ref{fig:linear-gmm} shows scatter plots of the joint samples alongside marginal density estimates. In the linear Gaussian test (Fig.~\ref{fig:linear-gmm}, top row), \texttt{Blade} and \texttt{EKS}, as sampling methods,
are able to capture the posterior mean and variance. In contrast, optimization-based algorithms like SCG, DPG, and EnKG operate by minimizing a surrogate objective and thus return point estimates, failing to capture posterior spread/uncertainty. 
In the multimodal non-isotropic Gaussian mixture case (Fig.~\ref{fig:linear-gmm}, 2nd row), \texttt{Blade} matches the analytic posterior most faithfully, recovering both modes and their relative weights without spurious samples. \texttt{EKS} collapses to a single over-dispersed Gaussian, and \texttt{EKS w/ DP} recovers the two modes but scatters a few outliers.
The detailed setup, derivations of the analytic posteriors, and additional evaluation metrics, are in Appendix~\ref{sec:exp-linear-gmm-setup}, with further quantitative results in Appendix~\ref{sec:additional-results}.

\subsection{Navier-Stokes equation}\label{sec:ns-experiment}
To test \texttt{Blade} in a higher-dimensional and more challenging setting, we consider the problem of recovering the initial vorticity field in the two-dimensional Navier–Stokes equations from partial, noisy observations taken at a later time. We pick this problem because the dynamics is highly nonlinear and mirrors many practical challenges in science and engineering including weather data assimilation~\citep{white2000ifs}, geophysics~\citep{liu2008simultaneous}, and fluid reconstruction~\citep{elsinga2006tomographic}. More importantly, this setting has standard probabilistic verifications available to test the quality of posterior samples.

\subsection{Ablation studies}\label{sec:ablation-study}
\begin{figure}[t]
    \centering
    \includegraphics[width=0.75\linewidth]{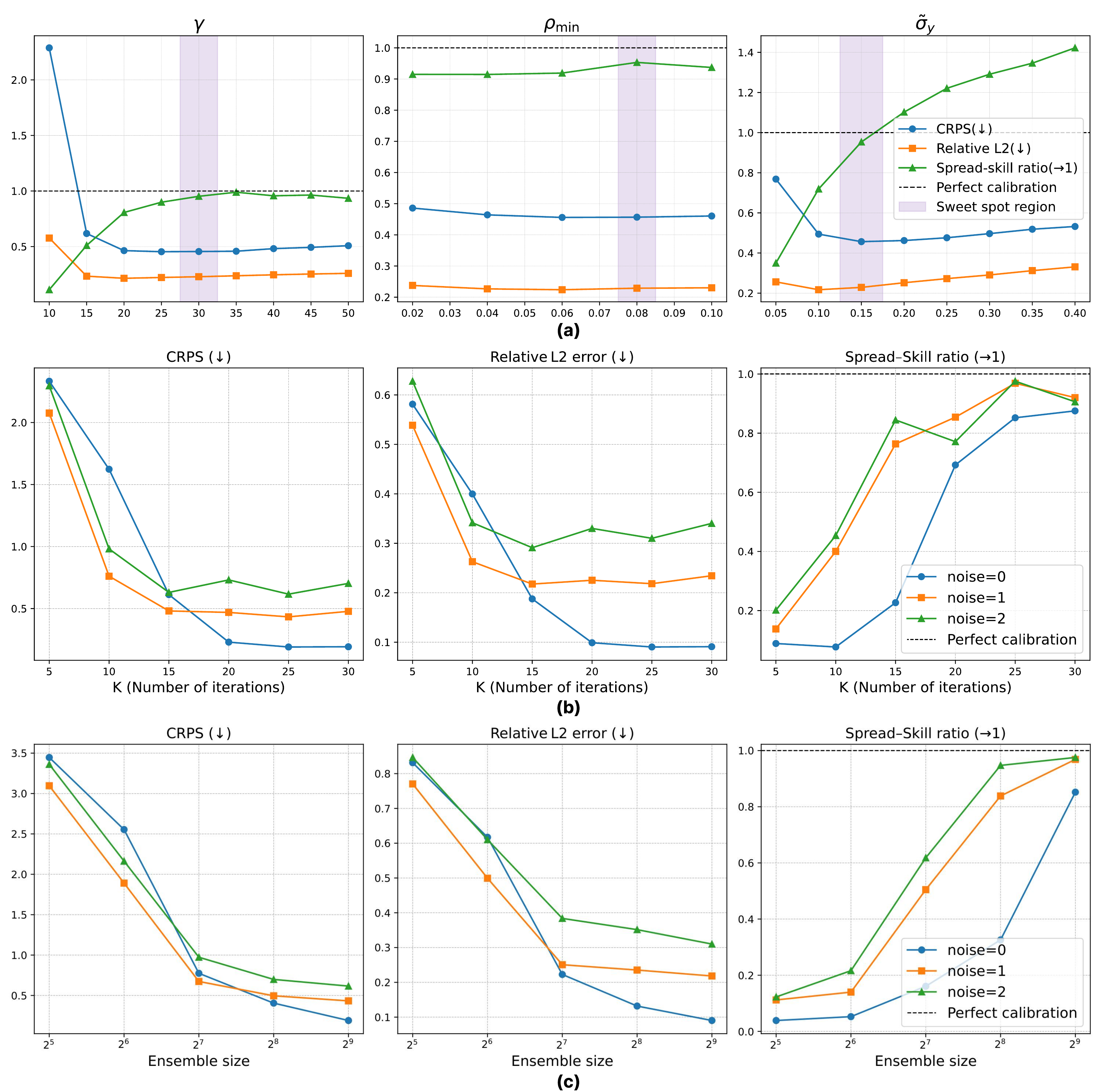}
    \caption{\texttt{Blade} improves predictably with more iterations and larger ensembles, and is robust across a broad range of hyperparameters. All panels report CRPS and relative L2 error (lower is better) and the spread-skill ratio (SSR), which equals 1 at perfect calibration. \textbf{(a)} Sensitivity to the three main hyperparameters $\gamma$ (discretization step scale), $\rho_{\mathrm{min}}$ (minimum coupling strength), and $\Tilde{\sigma}_{\rvy}$ (likelihood–spread factor); accuracy and calibration remain near-optimal across the shaded sweet-spot bands. \textbf{(b)} Increasing the number of split-Gibbs iterations $K$ improves accuracy and calibration together. \textbf{(c)} Accuracy and calibration improve with ensemble size, motivating our default of 512. Panels (b,c) are shown at three observation-noise levels.}
    \label{fig:ablation-study}
\end{figure}
\vspace{-5pt}

\textbf{Problem setup.} We take the Navier-Stokes problem formulation in InverseBench~\citep{zhenginversebench} where a non-trivial distribution is considered. The initial vorticity $\rvx^*$ in resolution $128\times 128$ is evolved forward with a numerical solver, then subsampled and corrupted with Gaussian noise of standard deviation $\sigma_{\mathrm{noise}}=0,1,2$. The observation $\rvy$ thus constitute a partial, noisy snapshot of the flow field. We use the publicly released dataset and pretrained diffusion prior from InverseBench. All the experiments are conducted on single GH200 GPU. Details are in Appendix~\ref{sec:navier-stokes-setup}; \texttt{Blade}'s hyperparameter settings for each noise level are listed in Table~\ref{tab:algo-hyperparameters}.

\textbf{Baselines.} We compare our algorithm against two classes of methods. 
The first class only requires a diffusion prior trained on unpaired data, including DPG~\citep{tang2024solving}, SCG~\citep{huang2024symbolic}, EnKG~\citep{zheng2024ensemble}, EKI~\citep{iglesias2013ensemble}, and FKD~\citep{singhal2025general,zhao2025conditional}. Since EKS~\citep{garbuno2020interacting} is defined only for Gaussian priors, we include two straightforward ways to incorporate a diffusion prior into it---\emph{EKS + DM}, which initializes the ensemble from diffusion samples, and \emph{EKS w/ DP}, which uses the diffusion score as the prior gradient---along with a localized variant~\citep{reich2021fokker,wagner2022ensemble}. The second class is provided as reference points, which requires additional training on paired data, including conditional diffusion model (CDM) and an end-to-end U-Net. The conditional diffusion learns the posterior distribution through conditional score matching. 
The U-Net directly learns to predict ground truth from the observation. 
For each noise regime we retrain both the conditional diffusion model and the U-Net from scratch, using the same training configuration. The details are in Appendix~\ref{sec:baseline-details}. 
To ensure each baseline is sufficiently tuned, we follow the two-stage hyperparameter-selection protocol of InverseBench~\citep{zhenginversebench}: a coarse grid search to narrow the search space, then Bayesian optimization on the validation set.

\textbf{Evaluation metrics.}
We evaluate from both probabilistic and deterministic perspectives~\citep{zhenginversebench, rasp2024weatherbench}. The three scalar metrics are the continuous ranked probability score (CRPS), spread-skill ratio (SSR), and relative L2 error (Rel L2 error): CRPS rewards both sharp and well-calibrated predictions, SSR diagnoses calibration alone (near one is desirable, and must be read alongside the error metrics), and Rel L2 measures point accuracy. Beyond these scalar summaries, we use the rank histogram as a finer-grained calibration diagnostic, resolving the shape of miscalibration---under-dispersion, over-dispersion, or bias---that SSR alone cannot. 
Formal definitions and implementation details of these metrics are in Appendix~\ref{sec:metrics}.  

\textbf{Results.} Table~\ref{tab:navier_stokes} summarizes performance under three observation noise levels. 
\texttt{Blade (main)} offers the best calibrated ensemble predictions: its CRPS is the best among all, and its SSR remains close to one. The other competing methods are too confident (SSR $<0.2$) and their predictions do not represent the true uncertainty. The CRPS of CDM has a very high CRPS despite SSR near one, which means that it produces overly diffuse distribution with large errors. 
\texttt{Blade(diag)} delivers the best point estimates with the lowest relative L2 but is under-dispersive with larger CRPS and small SSR compared to \texttt{Blade (main)}, confirming our theoretical insights in Sec.\ref{sec:practical-algo}. Overall, \texttt{Blade (main)} achieves both accurate recovery and reliable uncertainty calibration. Fig.~\ref{fig:linear-gmm}(b) shows qualitative samples and rank histograms: only \texttt{Blade}'s histogram is near-uniform, while the derivative-free baselines are under-dispersed and the conditional DM is over-dispersed. \texttt{EKS w/ DP} reads the learned score at small noise (Appendix~\ref{sec:baseline-details}), where it is least reliable; \texttt{Blade} only samples the prior and is more stable under score error. 
The runtime comparison is reported in Fig.~\ref{fig:runtime-comparison}. As shown, \texttt{Blade} runs slower than EKS but faster than the other derivative-free baselines. Moreover, its inherently parallel design allows for straightforward scaling, similar to common data parallelism.



\textbf{Impact of different hyperparameters.}
We perform comprehensive controlled experiments to understand the impact of different hyperparameters and design choices.  We sweep the three main hyperparameters of \texttt{Blade} (Fig.~\ref{fig:ablation-study}a). Formal definitions of hyperparameters, detailed discussion, and additional results are in Appendix~\ref{sec:implementation-details}. In a nutshell, we observe that the discretization step scale $\gamma$ and the minimum coupling strength $\rho_{\mathrm{min}}$ both have a broad plateau where both accuracy and calibration remain near‐optimal, simplifying practical tuning.  $\Tilde{\sigma}_{\rvy}$ is the most important factor that affects the performance. Increasing $\Tilde{\sigma}_{y}$ widens the ensemble as expected. A sweet spot region is around $0.15$.

\textbf{Test-time scaling.} We vary the number of split-Gibbs alternations $K$ and track the performance of \texttt{Blade} across different observation noise levels.  
Predictive accuracy and calibration both improve rapidly up to $K=20$ and plateau beyond it, so extra iterations yield diminishing returns (Fig.~\ref{fig:ablation-study}b). 
This trend aligns with the convergence and stability results in Eq.~\eqref{eq:convergence-thm}.

\textbf{Ensemble size.} Across all noise levels, all metrics keep improving as the ensemble grows (Fig.~\ref{fig:ablation-study}c). We fix it at 512 by default, balancing quality against the per-step cost that scales with ensemble size.

\textbf{Data scaling.} We evaluate how the performance of \texttt{Blade} depends on the pre-trained diffusion prior. We retrain the diffusion prior with progressively larger data subsets (ranging from $10\times 2^7$ to $10\times 2^{11}$).  We compare the resulting performance to that of the end-to-end U-Net and CDM, both retrained for each subset and each noise level. As shown in Fig.~\ref{fig:data-scaling}, \texttt{Blade}  surpasses the baselines by a significant margin. In contrast, the U-Net improves only modestly with data size. These results indicate that \texttt{Blade} is sample-efficient, can efficiently exploit extra prior data when available yet maintaining decent performance even in low-data regime.

\textbf{Additional ablation studies.}
We conduct additional ablation studies of \texttt{Blade} in Appendix~\ref{sec:implementation-details}, including the design choice of the annealing schedule for the coupling strength, the impact of initialization, and the effect of resample strategy. 
\begin{table}[t]
        \centering
        \caption{Quantitative evaluation on FFHQ 256x256 dataset. We report average metrics for image quality and samples consistency on three tasks. Measurement noise level $\sigma=0.05$ is used if not otherwise stated. ($\dagger$: the general PnP-DM algorithm that uses Langevin dynamics for likelihood step.)}
        \label{tab:ffhq256}
        \resizebox{0.95\linewidth}{!}{
        {\small
        \begin{tabular}{lcccccccccccc}\toprule
    & \multicolumn{3}{c}{\textbf{Inpaint (box)}} & \multicolumn{3}{c}{\textbf{SR ($\times$4)}} & \multicolumn{3}{c}{\textbf{Phase retrieval}}
    \\
    \cmidrule(lr){2-4}\cmidrule(lr){5-7} \cmidrule(lr){8-10} \cmidrule(lr){11-13}
               & PSNR$\uparrow$ & SSIM$\uparrow$& LPIPS$\downarrow$   & PSNR$\uparrow$  & SSIM$\uparrow$ & LPIPS$\downarrow$ & PSNR$\uparrow$  & SSIM$\uparrow$ & LPIPS$\downarrow$\\ \midrule
    \textbf{Black-box access} &  \\
    FKD~\citep{singhal2025general} & 17.62 & 0.628 & 0.345 & 24.07 & 0.778 & 0.215 & 11.22 & 0.418 & 0.584 \\
    DPG~\citep{tang2024solving} & 20.89 & 0.752 & 0.184 & 28.12 & 0.831 & 0.126 & 8.76 & 0.297 & 0.663 \\
    EnKG~\citep{zheng2024ensemble} & 21.70 & 0.727 & 0.286 & 27.17 & 0.773 & 0.237 & 24.02 & \underline{0.796} & \underline{0.232} \\
    \texttt{Blade (diag)} & 20.24 & 0.804 & 0.191 & 29.43 & 0.853 & \underline{0.098} & \underline{24.55} & 0.786 & 0.263 \\
    \texttt{Blade (main)} & \underline{24.07} & \textbf{0.861} & \underline{0.144} & \underline{31.24} & \textbf{0.900} & \underline{0.098} & 23.38 & 0.760 & 0.266 \\ \midrule
    \textcolor{gray}{\textbf{Gradient access}} &        \\
    \textcolor{gray}{DPS~\citep{chung2023diffusion}} & \textcolor{gray}{21.77} & \textcolor{gray}{0.767} & \textcolor{gray}{0.213} & \textcolor{gray}{24.90} & \textcolor{gray}{0.710} & \textcolor{gray}{0.265} & \textcolor{gray}{16.79} & \textcolor{gray}{0.589} & \textcolor{gray}{0.448} \\
    \textcolor{gray}{PnP-DM$^\dagger$~\citep{wu2024principled}} & \textcolor{gray}{22.17} & \textcolor{gray}{\underline{0.832}} & \textcolor{gray}{\underline{0.136}} & \textcolor{gray}{25.86} & \textcolor{gray}{0.808} & \textcolor{gray}{0.193} & \textcolor{gray}{18.98} & \textcolor{gray}{0.650} & \textcolor{gray}{0.409} \\
    \textcolor{gray}{DAPS~\citep{zhang2024improving}} & \textcolor{gray}{\textbf{25.61}} & \textcolor{gray}{\underline{0.846}} & \textcolor{gray}{\textbf{0.108}} & \textcolor{gray}{\textbf{31.65}} & \textcolor{gray}{\underline{0.881}} & \textcolor{gray}{\textbf{0.075}} & \textcolor{gray}{\textbf{25.26}} & \textcolor{gray}{\textbf{0.798}} & \textcolor{gray}{\textbf{0.202}} \\
    \bottomrule
    \end{tabular}}}
\end{table}

\begin{table}[t]
    \centering
    \caption{Evaluating \texttt{Blade} on the black hole imaging benchmark from InverseBench~\citep{zhenginversebench}. Note that this benchmark only uses deterministic metrics with a focus on accurate point estimate. \texttt{Blade}'s goal is well-calibrated posterior sampling. We therefore regard the results as complementary evidence.}
    \label{tab:black-hole-imaging}
    \resizebox{0.7\linewidth}{!}{
    \small
    \begin{tabular}{lcccc}
    \toprule
    \textbf{Method} & \textbf{PSNR}$\uparrow$ & \textbf{Blur PSNR}$\uparrow$ & $\chi^2_{cp} \rightarrow 1$ & $\chi^2_{camp} \rightarrow 1$ \\
    \midrule
    \textbf{Black-box access} \\
    \texttt{Blade(diag)}  & \underline{30.83} & \underline{36.22} & 2.12 & 1.61 \\
    \texttt{Blade(main)}  & \textbf{31.30} & \textbf{36.81} & 2.24 & 1.95 \\
    \midrule
    \textcolor{gray}{\textbf{Gradient access}} \\
    \textcolor{gray}{DPS~\citep{chung2023diffusion}}      & \textcolor{gray}{25.86} & \textcolor{gray}{32.94} & \textcolor{gray}{8.76} & \textcolor{gray}{5.46} \\
    \textcolor{gray}{LGD~\citep{song2023lossguided}}      & \textcolor{gray}{21.22} & \textcolor{gray}{26.06} & \textcolor{gray}{13.24} & \textcolor{gray}{13.22} \\
    \textcolor{gray}{PnPDM~\citep{wu2024principled}}    & \textcolor{gray}{26.07} & \textcolor{gray}{32.88} & \textcolor{gray}{\underline{1.31}} & \textcolor{gray}{\textbf{1.20}} \\
    \textcolor{gray}{DAPS~\citep{zhang2024improving}}     & \textcolor{gray}{25.60} & \textcolor{gray}{32.78} & \textcolor{gray}{\textbf{1.30}} & \textcolor{gray}{\underline{1.23}} \\
    \textcolor{gray}{RED-Diff~\citep{mardani2024a}} & \textcolor{gray}{23.77} & \textcolor{gray}{29.13} & \textcolor{gray}{1.85} & \textcolor{gray}{2.05} \\
    \textcolor{gray}{DiffPIR~\citep{zhu2023denoising}}  & \textcolor{gray}{25.01} & \textcolor{gray}{31.86} & \textcolor{gray}{3.27} & \textcolor{gray}{2.97} \\
    \bottomrule
    \end{tabular}}
    \end{table}

\subsection{Breadth across forward models}\label{sec:exp-on-point-estimate}
To show the breadth of \texttt{Blade}, we evaluate it on benchmarks with various different forward models: image restoration (linear and nonlinear tasks) and black hole imaging. 
These benchmarks assume a single ground truth and report only deterministic, point-estimate metrics, so they probe accuracy rather than the posterior calibration that is our main focus. 
We use 512 particles for black hole imaging and 1024 for image restoration, where the larger ensemble brings a marked improvement.

\textbf{Image restoration.} We test \texttt{Blade} on three image restoration tasks on the FFHQ256 dataset: box inpainting, $4\times$ super-resolution, and phase retrieval (nonlinear). We compare it against black-box and gradient-based methods (Table~\ref{tab:ffhq256}). \texttt{Blade} is the best black-box method on the linear tasks and, despite using only forward evaluations, is competitive with DAPS~\citep{zhang2024improving}, a state-of-the-art gradient-based restoration method---even surpassing it in SSIM on both linear tasks, though DAPS keeps an edge on PSNR and LPIPS. On the highly ill-posed, nonlinear phase-retrieval task \texttt{Blade} stays robust (PSNR $\approx 24.5$), on par with EnKG and far above DPG and FKD, which collapse. Problem setups and evaluation metrics are in Appendix~\ref{sec:image-appendix}.

\textbf{Black hole imaging.} Following InverseBench~\citep{zhenginversebench}, this task reconstructs a black hole image from the sparse, noisy measurements of the Event Horizon Telescope, with a highly non-convex forward model built from closure quantities (Appendix~\ref{sec:bh-appendix}). As shown in Table~\ref{tab:black-hole-imaging}, \texttt{Blade} is the only derivative-free method, yet it attains the best PSNR and Blur PSNR by a clear margin ($\approx$4 dB over the gradient-based method); its EHT closure statistics ($\chi^2_{cp}$, $\chi^2_{camp}$) stay competitive, though the gradient-based DAPS and PnP-DM match the $\chi^2\!\to\!1$ target more closely.




\section{Conclusion}

\texttt{Blade} brings well-calibrated posterior samples to derivative-free Bayesian inversion on high-dimensional, nonlinear problems. 
Optimization-based solvers collapse the posterior spread and prior derivative-free samplers stay miscalibrated here. 
\texttt{Blade} closes the gap with an interacting-particle scheme in the split Gibbs framework, alternating a derivative-free ensemble likelihood step with a diffusion prior step. This makes it a promising fit for problems like weather data assimilation, where calibrated uncertainty is essential.

\texttt{Blade} also scales reliably with more iterations and particles, shown in our convergence and stability analysis and empirical experiments. 
Evolving a large ensemble in parallel is its main cost, but that is exactly the workload modern distributed hardware handles well, the same parallel scaling behind today's large language models. 
We therefore see \texttt{Blade} as a promising derivative-free candidate for large-scale Bayesian inversion.

One limitation is that probabilistic benchmarks for high-dimensional nonlinear inverse problems remain scarce in the field, so we verify calibration only on 2D Navier-Stokes. 
Our image-restoration and black-hole-imaging benchmarks broaden the forward models but report only point-estimate metrics. 
A natural next step is to build one closer to numerical weather prediction---3D, with a more realistic dynamical core and a larger diffusion prior---moving \texttt{Blade} toward operational systems such as the ECMWF Integrated Forecasting System~\citep{white2000ifs}.



\bibliography{references}

@inproceedings{
mardani2024a,
title={A Variational Perspective on Solving Inverse Problems with Diffusion Models},
author={Morteza Mardani and Jiaming Song and Jan Kautz and Arash Vahdat},
booktitle={The Twelfth International Conference on Learning Representations},
year={2024},
url={https://openreview.net/forum?id=1YO4EE3SPB}
}

@article{chada2022convergence,
  title={Convergence acceleration of ensemble Kalman inversion in nonlinear settings},
  author={Chada, Neil and Tong, Xin},
  journal={Mathematics of Computation},
  volume={91},
  number={335},
  pages={1247--1280},
  year={2022}
}

@article{singhal2025general,
  title={A general framework for inference-time scaling and steering of diffusion models},
  author={Singhal, Raghav and Horvitz, Zachary and Teehan, Ryan and Ren, Mengye and Yu, Zhou and McKeown, Kathleen and Ranganath, Rajesh},
  journal={arXiv preprint arXiv:2501.06848},
  year={2025}
}

@article{wagner2022ensemble,
  title={The ensemble Kalman filter for rare event estimation},
  author={Wagner, Fabian and Papaioannou, Iason and Ullmann, Elisabeth},
  journal={SIAM/ASA Journal on Uncertainty Quantification},
  volume={10},
  number={1},
  pages={317--349},
  year={2022},
  publisher={SIAM}
}

@article{zhao2025conditional,
  title={Conditional sampling within generative diffusion models},
  author={Zhao, Zheng and Luo, Ziwei and Sj{\"o}lund, Jens and Sch{\"o}n, Thomas},
  journal={Philosophical Transactions A},
  volume={383},
  number={2299},
  pages={20240329},
  year={2025},
  publisher={The Royal Society}
}

@article{reich2021fokker,
  title={Fokker--Planck particle systems for Bayesian inference: Computational approaches},
  author={Reich, Sebastian and Weissmann, Simon},
  journal={SIAM/ASA Journal on Uncertainty Quantification},
  volume={9},
  number={2},
  pages={446--482},
  year={2021},
  publisher={SIAM}
}

@inproceedings{achituveinverse,
  title={Inverse Problem Sampling in Latent Space Using Sequential Monte Carlo},
  author={Achituve, Idan and Habi, Hai Victor and Rosenfeld, Amir and Netzer, Arnon and Diamant, Idit and Fetaya, Ethan},
  booktitle={Forty-second International Conference on Machine Learning},
  year={2025}
}

@inproceedings{janati2025mixture,
  title={A Mixture-Based Framework for Guiding Diffusion Models},
  author={Janati, Yazid and Moufad, Badr and Abou El Qassime, Mehdi and Durmus, Alain Oliviero and Moulines, Eric and Olsson, Jimmy},
  booktitle={Forty-second International Conference on Machine Learning},
  year={2025}
}

@misc{arjovsky2017principledmethodstraininggenerative,
      title={Towards Principled Methods for Training Generative Adversarial Networks}, 
      author={Martin Arjovsky and Léon Bottou},
      year={2017},
      eprint={1701.04862},
      archivePrefix={arXiv},
      primaryClass={stat.ML},
      url={https://arxiv.org/abs/1701.04862}, 
}

@inproceedings{rombach2022high,
  title={High-resolution image synthesis with latent diffusion models},
  author={Rombach, Robin and Blattmann, Andreas and Lorenz, Dominik and Esser, Patrick and Ommer, Bj{\"o}rn},
  booktitle={Proceedings of the IEEE/CVF conference on computer vision and pattern recognition},
  pages={10684--10695},
  year={2022}
}

@article{li2020fourier,
  title={Fourier neural operator for parametric partial differential equations},
  author={Li, Zongyi and Kovachki, Nikola and Azizzadenesheli, Kamyar and Liu, Burigede and Bhattacharya, Kaushik and Stuart, Andrew and Anandkumar, Anima},
  journal={arXiv preprint arXiv:2010.08895},
  year={2020}
}

@article{takamoto2022pdebench,
  title={Pdebench: An extensive benchmark for scientific machine learning},
  author={Takamoto, Makoto and Praditia, Timothy and Leiteritz, Raphael and MacKinlay, Daniel and Alesiani, Francesco and Pfl{\"u}ger, Dirk and Niepert, Mathias},
  journal={Advances in Neural Information Processing Systems},
  volume={35},
  pages={1596--1611},
  year={2022}
}

@article{anderson1996method,
  title={A method for producing and evaluating probabilistic forecasts from ensemble model integrations},
  author={Anderson, Jeffrey L},
  journal={Journal of climate},
  volume={9},
  number={7},
  pages={1518--1530},
  year={1996}
}

@article{hamill1997reliability,
  title={Reliability diagrams for multicategory probabilistic forecasts},
  author={Hamill, Thomas M},
  journal={Weather and forecasting},
  volume={12},
  number={4},
  pages={736--741},
  year={1997}
}

@inproceedings{talagrand1999evaluation,
  title={Evaluation of probabilistic prediction systems},
  author={Talagrand, Olivier},
  booktitle={Workshop Proceedings" Workshop on Predictability", 20-22 October 1997, ECMWF, Reading, UK},
  year={1999}
}

@article{elsinga2006tomographic,
  title={Tomographic particle image velocimetry},
  author={Elsinga, Gerrit E and Scarano, Fulvio and Wieneke, Bernhard and van Oudheusden, Bas W},
  journal={Experiments in fluids},
  volume={41},
  number={6},
  pages={933--947},
  year={2006},
  publisher={Springer}
}

@article{liu2008simultaneous,
  title={Simultaneous inversion of mantle properties and initial conditions using an adjoint of mantle convection},
  author={Liu, Lijun and Gurnis, Michael},
  journal={Journal of Geophysical Research: Solid Earth},
  volume={113},
  number={B8},
  year={2008},
  publisher={Wiley Online Library}
}

@article{he2007stability,
  title={Stability and convergence of the Crank--Nicolson/Adams--Bashforth scheme for the time-dependent Navier--Stokes equations},
  author={He, Yinnian and Sun, Weiwei},
  journal={SIAM Journal on Numerical Analysis},
  volume={45},
  number={2},
  pages={837--869},
  year={2007},
  publisher={SIAM}
}

@article{zhang2022fast,
  title={Fast sampling of diffusion models with exponential integrator},
  author={Zhang, Qinsheng and Chen, Yongxin},
  journal={arXiv preprint arXiv:2204.13902},
  year={2022}
}

@article{lee2022convergence,
  title={Convergence for score-based generative modeling with polynomial complexity},
  author={Lee, Holden and Lu, Jianfeng and Tan, Yixin},
  journal={Advances in Neural Information Processing Systems},
  volume={35},
  pages={22870--22882},
  year={2022}
}

@article{chen2022sampling,
  title={Sampling is as easy as learning the score: theory for diffusion models with minimal data assumptions},
  author={Chen, Sitan and Chewi, Sinho and Li, Jerry and Li, Yuanzhi and Salim, Adil and Zhang, Anru R},
  journal={arXiv preprint arXiv:2209.11215},
  year={2022}
}

@article{garbuno2020affine,
  title={Affine invariant interacting Langevin dynamics for Bayesian inference},
  author={Garbuno-Inigo, Alfredo and Nusken, Nikolas and Reich, Sebastian},
  journal={SIAM Journal on Applied Dynamical Systems},
  volume={19},
  number={3},
  pages={1633--1658},
  year={2020},
  publisher={SIAM}
}

@article{fortin2014should,
  title={Why should ensemble spread match the RMSE of the ensemble mean?},
  author={Fortin, Vincent and Abaza, Mabrouk and Anctil, Francois and Turcotte, Raphael},
  journal={Journal of Hydrometeorology},
  volume={15},
  number={4},
  pages={1708--1713},
  year={2014}
}

@article{wu2023practical,
  title={Practical and asymptotically exact conditional sampling in diffusion models},
  author={Wu, Luhuan and Trippe, Brian and Naesseth, Christian and Blei, David and Cunningham, John P},
  journal={Advances in Neural Information Processing Systems},
  volume={36},
  pages={31372--31403},
  year={2023}
}

@article{trippe2022diffusion,
  title={Diffusion probabilistic modeling of protein backbones in 3d for the motif-scaffolding problem},
  author={Trippe, Brian L and Yim, Jason and Tischer, Doug and Baker, David and Broderick, Tamara and Barzilay, Regina and Jaakkola, Tommi},
  journal={arXiv preprint arXiv:2206.04119},
  year={2022}
}

@book{sanz2023inverse,
  title={Inverse problems and data assimilation},
  author={Sanz-Alonso, Daniel and Stuart, Andrew and Taeb, Armeen},
  volume={107},
  year={2023},
  publisher={Cambridge University Press}
}

@book{white2000ifs,
  title={IFS documentation: Part III: Dynamics and numerical procedures (CY21r4)},
  author={White, Peter W},
  year={2000},
  publisher={European Centre for Medium-Range Weather Forecasts}
}

@book{park2013data,
  title={Data assimilation for atmospheric, oceanic and hydrologic applications (Vol. II)},
  author={Park, Seon Ki and Xu, Liang},
  year={2013},
  publisher={Springer}
}

@article{bannister2017review,
  title={A review of operational methods of variational and ensemble-variational data assimilation},
  author={Bannister, Ross N},
  journal={Quarterly Journal of the Royal Meteorological Society},
  volume={143},
  number={703},
  pages={607--633},
  year={2017},
  publisher={Wiley Online Library}
}

@book{hadamard2014lectures,
  title={Lectures on Cauchy's problem in linear partial differential equations},
  author={Hadamard, Jacques},
  year={2014},
  publisher={Courier Corporation}
}

@inproceedings{zhenginversebench,
  title={InverseBench: Benchmarking Plug-and-Play Diffusion Priors for Inverse Problems in Physical Sciences},
  author={Zheng, Hongkai and Chu, Wenda and Zhang, Bingliang and Wu, Zihui and Wang, Austin and Feng, Berthy and Zou, Caifeng and Sun, Yu and Kovachki, Nikola Borislavov and Ross, Zachary E and others},
  booktitle={The Thirteenth International Conference on Learning Representations},
  year = {2025}
}

@inproceedings{chu2025split,
  title={Split gibbs discrete diffusion posterior sampling},
  author={Chu, Wenda and Wu, Zihui and Chen, Yifan and Song, Yang and Yue, Yisong},
  booktitle={Advances in neural information processing systems},
  year={2025}
}

@article{zamo2018estimation,
  title={Estimation of the continuous ranked probability score with limited information and applications to ensemble weather forecasts},
  author={Zamo, Micha{\"e}l and Naveau, Philippe},
  journal={Mathematical Geosciences},
  volume={50},
  number={2},
  pages={209--234},
  year={2018},
  publisher={Springer}
}

@inproceedings{zhu2023denoising,
  title={Denoising diffusion models for plug-and-play image restoration},
  author={Zhu, Yuanzhi and Zhang, Kai and Liang, Jingyun and Cao, Jiezhang and Wen, Bihan and Timofte, Radu and Van Gool, Luc},
  booktitle={Proceedings of the IEEE/CVF Conference on Computer Vision and Pattern Recognition},
  pages={1219--1229},
  year={2023}
}

@article{rasp2024weatherbench,
  title={Weatherbench 2: A benchmark for the next generation of data-driven global weather models},
  author={Rasp, Stephan and Hoyer, Stephan and Merose, Alexander and Langmore, Ian and Battaglia, Peter and Russell, Tyler and Sanchez-Gonzalez, Alvaro and Yang, Vivian and Carver, Rob and Agrawal, Shreya and others},
  journal={Journal of Advances in Modeling Earth Systems},
  volume={16},
  number={6},
  pages={e2023MS004019},
  year={2024},
  publisher={Wiley Online Library}
}

@article{gneiting2007strictly,
  title={Strictly proper scoring rules, prediction, and estimation},
  author={Gneiting, Tilmann and Raftery, Adrian E},
  journal={Journal of the American statistical Association},
  volume={102},
  number={477},
  pages={359--378},
  year={2007},
  publisher={Taylor \& Francis}
}

@article{roberts1996exponential,
  title={Exponential convergence of Langevin distributions and their discrete approximations},
  author={Roberts, Gareth O and Tweedie, Richard L},
  year={1996}
}

@article{ma2015complete,
  title={A complete recipe for stochastic gradient MCMC},
  author={Ma, Yi-An and Chen, Tianqi and Fox, Emily},
  journal={Advances in neural information processing systems},
  volume={28},
  year={2015}
}

@article{nusken2019note,
  title={Note on interacting Langevin diffusions: gradient structure and ensemble Kalman sampler by Garbuno-Inigo, Hoffmann, Li and Stuart},
  author={N{\"u}sken, Nikolas and Reich, Sebastian},
  journal={arXiv preprint arXiv:1908.10890},
  year={2019}
}

@article{booton1954nonlinear,
  title={Nonlinear control systems with random inputs},
  author={Booton, Richard C},
  journal={IRE Transactions on Circuit Theory},
  volume={1},
  number={1},
  pages={9--18},
  year={1954},
  publisher={IEEE}
}

@article{kim2024derivative,
  title={Derivative-Free Diffusion Manifold-Constrained Gradient for Unified XAI},
  author={Kim, Won Jun and Chung, Hyungjin and Kim, Jaemin and Lee, Sangmin and Sim, Byeongsu and Ye, Jong Chul},
  journal={arXiv preprint arXiv:2411.15265},
  year={2024}
}

@article{evensen1994sequential,
  title={Sequential data assimilation with a nonlinear quasi-geostrophic model using Monte Carlo methods to forecast error statistics},
  author={Evensen, Geir},
  journal={Journal of Geophysical Research: Oceans},
  volume={99},
  number={C5},
  pages={10143--10162},
  year={1994},
  publisher={Wiley Online Library}
}

@article{iglesias2016regularizing,
  title={A regularizing iterative ensemble Kalman method for PDE-constrained inverse problems},
  author={Iglesias, Marco A},
  journal={Inverse Problems},
  volume={32},
  number={2},
  pages={025002},
  year={2016},
  publisher={IOP Publishing}
}

@article{chada2020tikhonov,
  title={Tikhonov regularization within ensemble Kalman inversion},
  author={Chada, Neil K and Stuart, Andrew M and Tong, Xin T},
  journal={SIAM Journal on Numerical Analysis},
  volume={58},
  number={2},
  pages={1263--1294},
  year={2020},
  publisher={SIAM}
}

@article{iglesias2013ensemble,
  title={Ensemble Kalman methods for inverse problems},
  author={Iglesias, Marco A and Law, Kody JH and Stuart, Andrew M},
  journal={Inverse Problems},
  volume={29},
  number={4},
  pages={045001},
  year={2013},
  publisher={IOP Publishing}
}

@inproceedings{
cardoso2024monte,
title={Monte Carlo guided Denoising Diffusion models for Bayesian linear inverse problems.},
author={Gabriel Cardoso and Yazid Janati el idrissi and Sylvain Le Corff and Eric Moulines},
booktitle={The Twelfth International Conference on Learning Representations},
year={2024},
url={https://openreview.net/forum?id=nHESwXvxWK}
}

@inproceedings{
dou2024diffusion,
title={Diffusion Posterior Sampling for Linear Inverse Problem Solving: A Filtering Perspective},
author={Zehao Dou and Yang Song},
booktitle={The Twelfth International Conference on Learning Representations},
year={2024},
url={https://openreview.net/forum?id=tplXNcHZs1}
}

@article{carrillo2022consensus,
  title={Consensus-based sampling},
  author={Carrillo, Jos{\'e} A and Hoffmann, Franca and Stuart, Andrew M and Vaes, Urbain},
  journal={Studies in Applied Mathematics},
  volume={148},
  number={3},
  pages={1069--1140},
  year={2022},
  publisher={Wiley Online Library}
}

@article{sun2024provable,
  title={Provable probabilistic imaging using score-based generative priors},
  author={Sun, Yu and Wu, Zihui and Chen, Yifan and Feng, Berthy T and Bouman, Katherine L},
  journal={IEEE Transactions on Computational Imaging},
  year={2024},
  publisher={IEEE}
}

@article{vempala2019rapid,
  title={Rapid convergence of the unadjusted langevin algorithm: Isoperimetry suffices},
  author={Vempala, Santosh and Wibisono, Andre},
  journal={Advances in neural information processing systems},
  volume={32},
  year={2019}
}

@article{karras2022elucidating,
  title={Elucidating the Design Space of Diffusion-Based Generative Models},
  author={Karras, Tero and Aittala, Miika and Aila, Timo and Laine, Samuli},
  journal={arXiv preprint arXiv:2206.00364},
  year={2022}
}

@inproceedings{
chung2023diffusion,
title={Diffusion Posterior Sampling for General Noisy Inverse Problems},
author={Hyungjin Chung and Jeongsol Kim and Michael Thompson Mccann and Marc Louis Klasky and Jong Chul Ye},
booktitle={The Eleventh International Conference on Learning Representations },
year={2023},
url={https://openreview.net/forum?id=OnD9zGAGT0k}
}

@article{zheng2024ensemble,
  title={Ensemble kalman diffusion guidance: A derivative-free method for inverse problems},
  author={Zheng, Hongkai and Chu, Wenda and Wang, Austin and Kovachki, Nikola and Baptista, Ricardo and Yue, Yisong},
  booktitle={Transactions on Machine Learning Research (TMLR)},
  year={2025}
}

@article{garbuno2020interacting,
  title={Interacting Langevin diffusions: Gradient structure and ensemble Kalman sampler},
  author={Garbuno-Inigo, Alfredo and Hoffmann, Franca and Li, Wuchen and Stuart, Andrew M},
  journal={SIAM Journal on Applied Dynamical Systems},
  volume={19},
  number={1},
  pages={412--441},
  year={2020},
  publisher={SIAM}
}

@article{wu2024principled,
  title={Principled Probabilistic Imaging using Diffusion Models as Plug-and-Play Priors},
  author={Wu, Zihui and Sun, Yu and Chen, Yifan and Zhang, Bingliang and Yue, Yisong and Bouman, Katherine L},
  journal={arXiv preprint arXiv:2405.18782},
  year={2024}
}

@article{kovachki2018eki,
  author    = {Kovachki, Nikola and Stuart, Andrew},
  title     = {Ensemble Kalman Inversion: A Derivative-Free Technique For Machine Learning Tasks},
  journal   = {arXiv preprint},
  year      = {2018},
  eprint    = {1808.03620},
  url       = {https://arxiv.org/abs/1808.03620},
}

@inproceedings{song2023lossguided,
  title         = {Loss-Guided Diffusion Models for Plug-and-Play Controllable Generation},
  author        = {Jiaming Song and Qinsheng Zhang and Hongxu Yin and Morteza Mardani and Ming-Yu Liu and Jan Kautz and Yongxin Chen and Arash Vahdat},
  booktitle     = {International Conference on Machine Learning (ICML)},
  month         = {July},
  year          = {2023},
}

@booktitle{zhang2024improving,
  title={Improving Diffusion Inverse Problem Solving with Decoupled Noise Annealing},
  author={Zhang, Bingliang and Chu, Wenda and Berner, Julius and Meng, Chenlin and Anandkumar, Anima and Song, Yang},
  booktitle = {IEEE/CVF Conference on Computer Vision and Pattern Recognition (CVPR)},
  year={2025}
}

@article{vono2019split,
  title={Split-and-augmented Gibbs sampler—Application to large-scale inference problems},
  author={Vono, Maxime and Dobigeon, Nicolas and Chainais, Pierre},
  journal={IEEE Transactions on Signal Processing},
  volume={67},
  number={6},
  pages={1648--1661},
  year={2019},
  publisher={IEEE}
}

@inproceedings{tang2024solving,
  title={Solving General Noisy Inverse Problem via Posterior Sampling: A Policy Gradient Viewpoint},
  author={Tang, Haoyue and Xie, Tian and Feng, Aosong and Wang, Hanyu and Zhang, Chenyang and Bai, Yang},
  booktitle={International Conference on Artificial Intelligence and Statistics},
  pages={2116--2124},
  year={2024},
  organization={PMLR}
}

@article{huang2024symbolic,
  title={Symbolic Music Generation with Non-Differentiable Rule Guided Diffusion},
  author={Huang, Yujia and Ghatare, Adishree and Liu, Yuanzhe and Hu, Ziniu and Zhang, Qinsheng and Sastry, Chandramouli S and Gururani, Siddharth and Oore, Sageev and Yue, Yisong},
  journal={arXiv preprint arXiv:2402.14285},
  year={2024}
}

@article{coeurdoux2023plug,
  title={Plug-and-play split Gibbs sampler: embedding deep generative priors in Bayesian inference},
  author={Coeurdoux, Florentin and Dobigeon, Nicolas and Chainais, Pierre},
  journal={arXiv preprint arXiv:2304.11134},
  year={2023}
}

@article{song2023solving,
  title={Solving inverse problems with latent diffusion models via hard data consistency},
  author={Song, Bowen and Kwon, Soo Min and Zhang, Zecheng and Hu, Xinyu and Qu, Qing and Shen, Liyue},
  journal={arXiv preprint arXiv:2307.08123},
  year={2023}
}

@article{huang2022iterated,
  title={Iterated Kalman methodology for inverse problems},
  author={Huang, Daniel Zhengyu and Schneider, Tapio and Stuart, Andrew M},
  journal={Journal of Computational Physics},
  volume={463},
  pages={111262},
  year={2022},
  publisher={Elsevier}
}

@article{cotter2013mcmc,
  title={MCMC methods for functions: modifying old algorithms to make them faster},
  author={Cotter, Simon L and Roberts, Gareth O and Stuart, Andrew M and White, David},
  year={2013}
}

@article{del2006sequential,
  title={Sequential monte carlo samplers},
  author={Del Moral, Pierre and Doucet, Arnaud and Jasra, Ajay},
  journal={Journal of the Royal Statistical Society Series B: Statistical Methodology},
  volume={68},
  number={3},
  pages={411--436},
  year={2006},
  publisher={Oxford University Press}
}

@article{gelman1997weak,
  title={Weak convergence and optimal scaling of random walk Metropolis algorithms},
  author={Gelman, Andrew and Gilks, Walter R and Roberts, Gareth O},
  journal={The annals of applied probability},
  volume={7},
  number={1},
  pages={110--120},
  year={1997},
  publisher={Institute of Mathematical Statistics}
}

@article{geyer1992practical,
  title={Practical markov chain monte carlo},
  author={Geyer, Charles J},
  journal={Statistical science},
  pages={473--483},
  year={1992},
  publisher={JSTOR}
}

@article{huang2022efficient,
  title={Efficient derivative-free Bayesian inference for large-scale inverse problems},
  author={Huang, Daniel Zhengyu and Huang, Jiaoyang and Reich, Sebastian and Stuart, Andrew M},
  journal={Inverse Problems},
  volume={38},
  number={12},
  pages={125006},
  year={2022},
  publisher={IOP Publishing}
}

@inproceedings{bouman2023generative,
  title={Generative plug and play: Posterior sampling for inverse problems},
  author={Bouman, Charles A and Buzzard, Gregery T},
  booktitle={2023 59th Annual Allerton Conference on Communication, Control, and Computing (Allerton)},
  pages={1--7},
  year={2023},
  organization={IEEE}
}

@article{wang2004image,
  title={Image quality assessment: from error visibility to structural similarity},
  author={Wang, Zhou and Bovik, Alan C and Sheikh, Hamid R and Simoncelli, Eero P},
  journal={IEEE transactions on image processing},
  volume={13},
  number={4},
  pages={600--612},
  year={2004},
  publisher={IEEE}
}

@inproceedings{zhang2018unreasonable,
  title={The unreasonable effectiveness of deep features as a perceptual metric},
  author={Zhang, Richard and Isola, Phillip and Efros, Alexei A and Shechtman, Eli and Wang, Oliver},
  booktitle={Proceedings of the IEEE conference on computer vision and pattern recognition},
  pages={586--595},
  year={2018}
}

@article{xu2024provably,
  title={Provably robust score-based diffusion posterior sampling for plug-and-play image reconstruction},
  author={Xu, Xingyu and Chi, Yuejie},
  journal={arXiv preprint arXiv:2403.17042},
  year={2024}
}
\bibliographystyle{paperstyle}

\appendix
\newpage

\section{Theory}
\subsection{Notation}\label{sec:notation}
We denote by $h(t)$ the drift coefficient in Eq.~\eqref{eq:reverse-diffusion} and $\delta(t)$ the diffusion coefficient: 
\begin{align}
    h(t) & \coloneq -\left(2\dot{\sigma}(t) \sigma(t) + \beta(t)\right) \label{eq:drift} \\
    \delta(t) & \coloneq \sqrt{2\dot{\sigma}(t) \sigma(t)} + \sqrt{2\beta(t)}. \label{eq:diffusion}
\end{align}

The Kullback–Leibler (KL) divergence between two distributions $\mu$ and $\Tilde{\mu}$ is $\KL$ defined by 
\begin{equation*}
    \KL(\mu || \Tilde{\mu}) = \int \mu \log \frac{\mu}{\Tilde{\mu}} = \EE_{\mu}\log \frac{\mu}{\Tilde{\mu}}.
\end{equation*}
The Fisher divergence between two distributions $\mu$ and $\Tilde{\mu}$ is $\FI$ defined by 
\begin{equation*}
    \FI(\mu || \Tilde{\mu}) = \int \mu \|\nabla \log \frac{\mu}{\Tilde{\mu}}\|_2^2 = \EE_{\mu} \left\|\nabla \log \frac{\mu}{\Tilde{\mu}}\right\|_2^2.
\end{equation*}
For a positive semi-definite matrix $B\in \RR^{n\times n}$, we denote by $\|\cdot\|_{B}$ the weighted norm defined by 
\begin{equation}\label{eq:def-weighted-norm}
    \|u\|^2_B = u^\top Bu, 
\end{equation}
where $u \in \RR^n$. For $\rvx \in \RR^n$, the divergence of a matrix $T(\rvx)\in\RR^{n\times n}$ is the vector field: 
\begin{equation}\label{eq:matrix-divergence}
    (\nabla_\rvx \cdot T)_i = \sum_{j=1}^n \frac{\partial T_{ij}}{\rvx_j}.
\end{equation}

\subsection{Proofs}
\label{sec:proofs}

\begin{assumption}\label{assumption:score}
    The average score approximation error of the diffusion model $s_\theta$ is bounded, 
    \begin{equation} \label{eq:score-bound}
         \epsilon_{\mathrm{score}} = \sup_{k=0,\dots,K-1} \left\{\frac{1}{t^*}\int_{T_k + t^\dagger}^{T_{k+1}} \frac{h(t)^2}{\delta(t)^2} \EE_{\mu_t} \| \nabla_{\rvx_t} \log p\left(\rvx_t;\sigma(t)\right) - s_{\theta}(\rvx_t,t) \|^2_2 \df t \right\} < +\infty, 
    \end{equation}
    where $h(t)$ is defined in Eq.~\eqref{eq:drift} and $\delta(t)$ is defined in Eq.~\eqref{eq:diffusion}.
\end{assumption}

\begin{assumption}\label{assumption:linearization}
    The average derivative approximation error of the linear surrogate model $A_t$ is bounded, 
    \begin{equation}\label{eq:linearization-bound}
        \epsilon_{\mathrm{model}} = \sup_{k=0,\dots,K-1} \left\{\frac{1}{t^\dagger}\int_{T_k}^{T_k + t^\dagger} \EE_{\mu_t} \left\| \nabla f(\rvz_t;\rvy) - \frac{1}{\sigma_y^2}A_t^\top (\mathcal{G}(\rvz_t^{(j)})-\rvy)\right\|^2_{C_t} \df t \right\} < +\infty,
    \end{equation}
    where $\|\cdot\|_{C_t}$ is the weighted norm defined in Eq.~\eqref{eq:def-weighted-norm}. 
\end{assumption}

\begin{assumption}\label{assumption-null-space-invariance}
    The Radon–Nikodym derivative $\frac{\df \mu_t}{\df \Tilde{\mu}_t}$ is constant along the null space of $C_t$ almost surely, where $C_t$ is the covariance matrix of $\Tilde{\mu}_t$.
\begin{equation*}
    \frac{\df \mu_t}{\df \Tilde{\mu}_t}(\rvx) = \frac{\df \mu_t}{\df \Tilde{\mu}_t}(\rvx + \rvv),\forall \rvv \in \mathrm{Ker}(C_t).
\end{equation*}
\end{assumption}

\begin{remark}
Assumption~\ref{assumption:score} coincides with the standard bounded score-matching error condition that underpins convergence results~\citep{lee2022convergence, chen2022sampling, wu2024principled}. 
The statistical linearization error $\epsilon_{\mathrm{model}}$ defined in Assumption~\ref{assumption:linearization} characterizes the L2 accuracy of the statistical linearization. Following the standard treatment~\citep{chada2022convergence}, it is uniformly bounded under the standard regularity assumptions on the forward model and a uniformly bounded ensemble covariance. 
We also make Assumption~\ref{assumption-null-space-invariance}, which is the weakest condition needed to bound the weighted Fisher divergence in our analysis. Two common sufficient
(but not necessary) scenarios are: (1) $C_t$ is full-rank; and (2) $\mu_t$ is absolutely continuous with respect to $\tilde{\mu}_t$ and both log densities are continuously differentiable.
\end{remark}

\begin{lemma}[Stationary distribution of the likelihood step]\label{lemma1}
    Assume the particle distribution is not a Dirac measure, 
    the dynamics of Eq.~\eqref{eq:cov-precond-langevin} admits $\pi^{Z|X=\rvx^{(j)}}(\rvz) \propto \exp(-f(\rvz;\rvy) - \frac{1}{2\rho^2}\|\rvz - \rvx^{(j)}\|_2^2)$ as a stationary distribution. Further, if the covariance matrix is positive definite, the stationary distribution is unique. 
\end{lemma}
\begin{proof}
    This result has been proved in various forms in the literature~\citep{ma2015complete, garbuno2020interacting}, we provide a simple proof of our use case for ease of understanding. Suppose $\mu_t(\rvz)$ is the probability density of $\rvz$ at time $t$. For the ease of notation, we ignore the particle index $j$ in $\rvz_t$. Let $\Phi(\rvz) = f(\rvz;\rvy)+ \frac{1}{2\rho^2}\|\rvz - \rvx^{(j)}\|_2^2$. The corresponding Fokker-Planck equation for Eq.~\eqref{eq:cov-precond-langevin} reads
    \begin{equation*}
        \frac{\partial \mu_t}{\partial t}= \nabla \cdot \left(\mu_t C_t \nabla \Phi(\rvz) \right) + \nabla\cdot\left( C_t\nabla \mu_t \right),
    \end{equation*}
    which can be rewritten as
    \begin{equation}\label{eq:fokker-planck}
        \frac{\partial \mu_t}{\partial t}= \nabla \cdot \left(\mu_t C_t (\nabla \Phi(\rvz) + \nabla \log\mu_t) \right).
    \end{equation}
    Let $\mu_{\infty}$ denote the stationary distribution of Eq.~\eqref{eq:fokker-planck}. We have
    \begin{equation*}
        0 = \nabla \cdot \left(\mu_t C_t (\nabla \Phi(\rvz) + \nabla \log\mu_{\infty}) \right).
    \end{equation*}
    If the particle distribution is not Dirac, $C_t\neq 0$ due to Lemma 2.1 in~\citet{garbuno2020interacting}. Therefore, 
    \[
    \nabla \Phi(\rvz) + \nabla \log\mu_{\infty}  = c,
    \]
    where $c$ is a constant. Integrating both sides gives
    \[
    \mu_{\infty}(\rvz) \propto \exp(-\Phi(\rvz)) = \exp(-f(\rvz;\rvy) - \frac{1}{2\rho^2}\|\rvz - \rvx^{(j)}\|_2^2),
    \]
    showing that $\pi^{Z|X=\rvx^{(j)}}(\rvz)$ is a stationary distribution of the dynamics of Eq.~\eqref{eq:cov-precond-langevin}. Further, if $C_t$ is positive definite, it ensures the irreducibility and strong Feller property, and the stationary distribution is unique~\citep{roberts1996exponential, ma2015complete}.
\end{proof}

\theoremStationary*

\begin{proof}
    We prove this by directly verifying the invariance property, i.e., if the samples $(\rvx, \rvz)$ are from the joint distribution $\pi^{XZ}$, then after one iteration of the algorithm, the new samples $(\rvx^\prime, \rvz^\prime)$ stay in the same distribution $\pi^{XZ}$. 
    By Lemma ~\ref{lemma1}, $\pi^{Z|X=\rvx}$ is a stationary distribution Eq.~\eqref{eq:cov-precond-langevin}. Therefore, after the oracle likelihood step, a stationary joint density of $(\rvx, \rvz^\prime)$ is given by 
    \begin{equation*}
        p(\rvx, \rvz^\prime)=\int \pi^{XZ}(\rvx,\rvz)\pi^{Z|X=\rvx}(\rvz^\prime) \df \rvz = \pi^X(\rvx)\pi^{Z|X=\rvx}(\rvz^\prime),
    \end{equation*}
    where $\pi^X$ is the marginal distribution. As shown in Eq.~\eqref{prior_posterior}, 
    after sampling $\rvx^\prime$ given $\rvz^\prime$ according to the prior step, the joint density of $(\rvx^\prime, \rvz^\prime)$ becomes 
    \begin{align*}
        p(\rvx^\prime, \rvz^\prime) & = \int p(\rvx, \rvz^\prime) \pi^{X|Z=\rvz^\prime}(\rvx^\prime)\df \rvx \\
        & = \int \pi^X(\rvx)\pi^{Z|X=\rvx}(\rvz^\prime) \pi^{X|Z=\rvz^\prime}(\rvx^\prime)\df \rvx \\
        & = \pi^{Z}(\rvz^\prime)\pi^{X|Z=\rvz^\prime}(\rvx^\prime) \\
        & = \pi^{XZ}(\rvx^\prime, \rvz^\prime),
    \end{align*}
    showing that the distribution of $(\rvx^\prime, \rvz^\prime)$ remains $\pi^{XZ}$ after one round of updates. Therefore, $\pi^{XZ}$ is a stationary distribution. Furthermore, by Lemma~\ref{lemma1}, if the particle covariance remains positive definite, the stationary distribution of the likelihood step is unique. Consequently, it follows that $\pi^{XZ}$ is the unique stationary distribution. 
\end{proof}

\begin{lemma} \label{lemma:matrix-ineq}
    Given the following pair of stochastic processes 
    \begin{align}
        \df \rvx_t & = b(\rvx_t,t)\df t + H(t)\df \rvw_t, \label{eq:lemma2-sde1} \\
        \df \Tilde{\rvx}_t & = \Tilde{b}(\Tilde{\rvx}_t, t) \df t + H(t) \df \rvw_t, \label{eq:lemma2-sde2}
    \end{align}
    where $b, \Tilde{b}: \RR^n\times \RR^+\rightarrow \RR^n$ are the drift terms, $H: \RR^+ \rightarrow \RR^n\times \RR^n$ is the diffusion term, $\rvw_t$ is the standard Wiener process. Let $\mu_t$ (respectively $\Tilde{\mu}_t$) be the law of $\rvx_t$ (respectively $\Tilde{\rvx}_t$), $C(t) \coloneq H(t)H(t)^\top$, and $\lambda^*_t$ be the smallest non-zero eigenvalue of $C(t)$. Assuming that $b_t - \Tilde{b}_t \in \mathrm{Range}(C(t))$ and Assumption~\ref{assumption-null-space-invariance} holds, we have
    \begin{equation}
        \frac{\partial}{\partial t}  \KL(\mu_t||\Tilde{\mu}_t)  \leq - \frac{\lambda^*_t}{4} \FI(\mu_t||\Tilde{\mu}_t) +  \EE_{\mu_t}\left\| b_t - \Tilde{b}_t\right\|^2_{C(t)^\dagger},
    \end{equation}
    where $C(t)^\dagger$ is the pseudo-inverse of $C(t)$.
\end{lemma}

\begin{proof}
    Since the diffusion terms only depend on $t$ and $C(t) = H(t)H(t)^\top$, the Fokker-Planck equations of Eq.~\eqref{eq:lemma2-sde1} and Eq.~\eqref{eq:lemma2-sde2} read
    \begin{align}
        \frac{\partial}{\partial t} \mu_t & = \nabla \cdot \left[ \left(\frac{1}{2}C(t)\nabla\log \mu_t - b_t\right)\mu_t \right],  \label{eq:lemma2-fp1}\\
        \frac{\partial}{\partial t} \Tilde{\mu}_t & = \nabla \cdot \left[ \left(\frac{1}{2}C(t)\nabla\log \Tilde{\mu}_t - \Tilde{b}_t\right) \Tilde{\mu}_t \right]. \label{eq:lemma2-fp2}
    \end{align}
    Let $r_t \coloneq \frac{\mu_t}{\Tilde{\mu}_t}$ and $\phi(r_t) \coloneq r_t \log r_t$ (so $\phi^\prime(r_t) = \frac{\df}{\df r_t}\phi(r_t)= \log r_t + 1$).  Differentiating the KL divergence gives 
    \begin{align}
        \frac{\partial}{\partial t} \KL(\mu_t||\Tilde{\mu}_t) & = \frac{\partial}{\partial t} \int \phi(r_t) \Tilde{\mu}_t \notag \\
        & = \int \left(\phi(r_t)\frac{\partial \Tilde{\mu}_t}{\partial t} + \phi^\prime(r_t) \frac{\partial r_t}{\partial t}   \Tilde{\mu}_t\right) \notag \\
        & = \int \left(\phi(r_t)\frac{\partial \Tilde{\mu}_t}{\partial t} + \phi^\prime(r_t) \frac{\partial \mu_t}{\partial t} - \phi^\prime(r_t) r_t \frac{\partial \Tilde{\mu}_t}{\partial t}\right) \notag \\
        & = \int\left((\log r_t +1) \frac{\partial \mu_t}{\partial t} - r_t \frac{\partial \Tilde{\mu}_t}{\partial t}\right), \label{eq:lemma2-kl-derivative}
    \end{align}
    where the last step uses the fact that $\phi(r_t)-r_t\phi^\prime(r_t)= -r_t$.
    Plugging Eq.~\eqref{eq:lemma2-fp1} and Eq.~\eqref{eq:lemma2-fp2} into Eq.~\eqref{eq:lemma2-kl-derivative} and applying integration by parts further gives 
    \begin{align}
        & \frac{\partial}{\partial t} \KL(\mu_t||\Tilde{\mu}_t)  \notag \\ 
        & = \int (\log r_t + 1)\nabla \cdot \left[ \left(\frac{1}{2}C(t)\nabla\log \mu_t - b_t\right)\mu_t \right] - \int r_t \nabla \cdot \left[ \left(\frac{1}{2}C(t)\nabla\log \Tilde{\mu}_t - \Tilde{b}_t\right) \Tilde{\mu}_t \right] \notag \\
        & = - \int \left\langle \nabla \log r_t, \frac{1}{2}C(t)\nabla\log \mu_t - b_t \right\rangle \mu_t+ \int \left\langle \nabla r_t, \frac{1}{2}C(t)\nabla\log \Tilde{\mu}_t - \Tilde{b}_t \right\rangle \Tilde{\mu}_t \notag \\
        & = - \int \left\langle \nabla \log r_t, \frac{1}{2}C(t)\nabla\log \mu_t - b_t \right\rangle \mu_t+ \int \left\langle \nabla \log r_t, \frac{1}{2}C(t)\nabla\log \Tilde{\mu}_t - \Tilde{b}_t \right\rangle \mu_t \notag \\
        & = - \int \left\langle \nabla \log r_t, \frac{1}{2}C(t) \left(\nabla\log \mu_t -  \nabla\log \Tilde{\mu}_t\right)\right\rangle \mu_t + \int \left\langle \nabla \log r_t, b_t - \Tilde{b}_t\right\rangle \mu_t \notag \\
        & = - \frac{1}{2} \int \left\langle \nabla \log r_t, C(t)\nabla \log r_t \right\rangle \mu_t + \int \left\langle \nabla \log r_t, b_t - \Tilde{b}_t\right\rangle \mu_t. \label{eq:lemma2-kl-derivative-2}
    \end{align}
    The weighted Young's inequality states that, for any $u,v\in\RR^n$, when  $v \in \mathrm{Range}(C)$, we have
    \begin{equation*}
        \left\langle u, v\right\rangle \leq \frac{1}{4} \left\langle u, Cu\right\rangle +  \left\langle v, C^\dagger v\right\rangle,
    \end{equation*}
    where $C^\dagger$ is the pseudo-inverse. By Assumption~\ref{assumption-null-space-invariance}, Eq.~\eqref{eq:lemma2-kl-derivative-2} can be bounded as follows
    \begin{align}
       & - \frac{1}{2} \int \left\langle \nabla \log r_t, C(t)\nabla\log r_t \right\rangle \mu_t + \int \left\langle \nabla \log r_t, b_t - \Tilde{b}_t\right\rangle \mu_t \notag \\
       & \leq - \frac{1}{4} \int \left\langle \nabla \log r_t, C(t)\nabla\log r_t \right\rangle \mu_t + \int \left\langle b_t - \Tilde{b}_t, C(t)^\dagger(b_t - \Tilde{b}_t)\right\rangle \mu_t \notag \\
       & \leq - \frac{\lambda^*_t}{4} \FI(\mu_t||\Tilde{\mu}_t) + \EE_{\mu_t}\left\| b_t - \Tilde{b}_t\right\|^2_{C(t)^\dagger}
    \end{align}
    where $\lambda^*_t$ is the smallest non-zero eigenvalue of $C(t)$. 
\end{proof}

\begin{remark}
    This is a generalization of Lemma A.4 in~\citet{wu2024principled} to the general matrix-valued diffusion term. Intuitively, the condition that $b_t - \Tilde{b}_t$ belongs to the range of $C(t)$ means that the two drift terms may only differ along the directions that are actually driven by noise. In the context of our proof below, this is always satisfied because the drift terms are either preconditioned with $C(t)$ or $C(t)$ is full-rank. 
\end{remark}

\theoremConvergence*

\begin{proof}
    For $t \in [T_k, T_k + t^\dagger], k=0,\dots, K-1$,  we apply Lemma~\ref{lemma:matrix-ineq} to the likelihood step with 
    \begin{align*}
        b(\rvz_t, t) & \coloneq - C_t\nabla f(\rvz_t;\rvy) - \frac{1}{\rho^2} C_t(\rvz_t-\rvx^{(j)}) \\  
        \Tilde{b}(\rvz_t, t) & \coloneq - C_t \frac{1}{\sigma_y^2}A_t^\top(\mathcal{G}(\rvz_t)-\rvy)  - \frac{1}{\rho^2} C_t(\rvz_t-\rvx^{(j)}) \\
        H(t) & = \sqrt{C_t}, 
    \end{align*}
    where $A_t = \EE_{\Tilde{\mu}_t}[ (\mathcal{G}(\rvz_t) - \EE_{q_t}\mathcal{G}(\rvz_t))\rvz_t^\top]  C_t^{-1}$ as defined in Eq.~\eqref{eq:linearization-solution}. Note that the condition $b-\Tilde{b} \in \mathrm{Range}(C_t)$ is satisfied as both drift terms are preconditioned with $C_t$.
    Thus, by Assumption~\ref{assumption-null-space-invariance}, we have
    \begin{align*}
        \frac{\partial}{\partial t}  \KL(\mu_t||\Tilde{\mu}_t) & \leq - \frac{\lambda^*_t}{4} \FI(\mu_t||\Tilde{\mu}_t) +  \EE_{\mu_t} \left\langle (b_t - \Tilde{b}_t), C_t^\dagger (b_t - \Tilde{b}_t)\right\rangle \\
        & \leq - \frac{\lambda^*}{4} \FI(\mu_t||\Tilde{\mu}_t) +  \EE_{\mu_t} \left\| \nabla f(\rvz_t;\rvy) - \frac{1}{\sigma_y^2} A_t^\top (\mathcal{G}(\rvz_t)-\rvy) \right\|^2_{C_t} 
    \end{align*}
    where $\lambda_t^*$ is the smallest non-zero eigenvalue of $C_t$ and $\lambda^*\coloneq\inf \lambda^*_t$. By Assumption~\ref{assumption:linearization} , integrating both sides over $[T_k, T_k + t^\dagger]$ gives 
    \begin{align}
         & \KL(\mu_{T_k + t^\dagger}||\Tilde{\mu}_{T_k+ t^\dagger}) - \KL(\mu_{T_k}||\Tilde{\mu}_{T_k})  \notag \\ 
         & \leq - \frac{\lambda^*_t}{4} \int_{T_k}^{T_k+t^\dagger}\FI(\mu_t||\Tilde{\mu}_t)\df t + \int_{T_k}^{T_k+t^\dagger}\EE_{\mu_t} \left\| \nabla f(\rvz_t;\rvy) - \frac{1}{\sigma_y^2}A_t^\top (\mathcal{G}(\rvz_t)-\rvy)\right\|^2_{C_t}\df t \notag \\
         & \leq - \frac{\lambda^*_t}{4} \int_{T_k}^{T_k+t^\dagger}\FI(\mu_t||\Tilde{\mu}_t)\df t + t^\dagger \epsilon_{\mathrm{model}}, \label{eq:ineq-linearization}
    \end{align}
    where $\epsilon_{\mathrm{model}}$ is defined in Eq.~\eqref{eq:linearization-bound}. For $t \in [T_k + t^\dagger , T_{k+1}], k=0,\dots, K-1$, we apply Lemma~\ref{lemma:matrix-ineq} to the prior step~\eqref{eq:reverse-diffusion} with 
    \begin{align*}
        b(\rvx_t, t) & \coloneq h(t) \nabla_{\rvx_t} \log p\left(\rvx_t;\sigma(t)\right) \\
        \Tilde{b}(\rvz_t, t) & \coloneq h(t) s_{\theta}(\rvx_t,t) \\
        H(t) & \coloneq \delta(t) \mI,
    \end{align*}
    where $h(t)$ is the drift coefficient defined in Eq.~\eqref{eq:drift}, $\delta(t)$ is the diffusion coefficient defined in Eq.~\eqref{eq:diffusion}, $s_{\theta}$ is the pre-trained diffusion model with score approximation error $\epsilon_{\mathrm{score}}$. Note that $H(t)H(t) ^\top$ is full-rank so that the condition of  Lemma~\ref{lemma:matrix-ineq} is satisfied. Therefore, we have 
    \begin{align*}
        \frac{\partial}{\partial t}  \KL(\mu_t||\Tilde{\mu}_t) & \leq - \frac{\delta(t)^2}{4}\FI(\mu_t||\Tilde{\mu}_t) +  \frac{h(t)^2}{\delta(t)^2} \EE_{\mu_t} \| \nabla_{\rvx_t} \log p\left(\rvx_t;\sigma(t)\right) - s_{\theta}(\rvx_t,t) \|_2^2  \\
        & \leq - \frac{\delta}{4} \FI(\mu_t||\Tilde{\mu}_t) + \frac{h(t)^2}{\delta(t)^2} \| \nabla_{\rvx_t} \log p\left(\rvx_t;\sigma(t)\right) - s_{\theta}(\rvx_t,t) \|^2_2, 
    \end{align*}
    where $\delta \coloneq \inf_{t\in[0,t^*]} \delta(t)^2$. Integrating both sides over $[T_k + t^\dagger , T_{k+1}]$ and applying Assumption~\ref{assumption:score} gives
    \begin{align}
        & \KL(\mu_{T_{k+1}}||\Tilde{\mu}_{T_{k+1}})  - \KL(\mu_{T_k + t^\dagger}||\Tilde{\mu}_{T_k+ t^\dagger})  \notag \\ 
        & \leq - \frac{\delta}{4} \int_{T_k + t^\dagger}^{T_{k+1}} \FI(\mu_t||\Tilde{\mu}_t)\df t + \int_{T_k + t^\dagger}^{T_{k+1}} \frac{h(t)^2}{\delta(t)^2}\EE_{\mu_t} \| \nabla_{\rvx_t} \log p\left(\rvx_t;\sigma(t)\right) - s_{\theta}(\rvx_t,t) \|_2^2 \df t \notag \\
        & \leq - \frac{\delta}{4} \int_{T_k + t^\dagger}^{T_{k+1}} \FI(\mu_t||\Tilde{\mu}_t)\df t + t^* \epsilon_{\mathrm{score}}, \label{eq:ineq-score}
    \end{align}
    where $\epsilon_{\mathrm{score}}$ is defined in Eq.~\eqref{eq:score-bound}. Summing up both sides of Eq.~\eqref{eq:ineq-linearization} and Eq.~\eqref{eq:ineq-score} for $k=0,\dots, K-1$ gives 
    \begin{equation*}
        \KL(\mu_{T_{K}}||\Tilde{\mu}_{T_{K}}) -  \KL(\mu_0||\Tilde{\mu}_0)  \leq - \frac{\min(\lambda^*, \delta)}{4} \int_0^{T_K} \FI(\mu_t||\Tilde{\mu}_t) \df t + K (t^\dagger\epsilon_{\mathrm{model}} + t^*\epsilon_{\mathrm{score}}). 
    \end{equation*}
    Rearranging the terms gives 
    \begin{align*}
        & \frac{1}{T_K} \int_0^{T_K} \FI(\mu_t||\Tilde{\mu}_t) \df t \\ 
        &  \leq \frac{4}{T_K\min(\lambda^*, \delta)}\left(\KL(\mu_0||\Tilde{\mu}_0) - \KL(\mu_{T_{K}}||\Tilde{\mu}_{T_{K}})\right) + \frac{4}{\min(\lambda^*, \delta)(t^\dagger +t^*)}(\epsilon_{\mathrm{model}} + \epsilon_{\mathrm{score}}) \\
        & \leq \frac{4}{\min(\lambda^*, \delta)} \left[\frac{\KL(\mu_0||\Tilde{\mu}_0) }{K(t^\dagger +t^*)} + \frac{t^\dagger\epsilon_{\mathrm{model}} + t^*\epsilon_{\mathrm{score}}}{t^\dagger +t^*} \right]. 
    \end{align*}
    Note that $\mu_0=\pi^X$ and we conclude the proof. 
\end{proof}

\begin{lemma}\label{lemma:finite-correction}
    Let $\pi(\rvz;\rvx^{(j)})$ denote the invariant measure associated with the potential $\Phi(\rvz; \rvx^{(j)})$ where $\nabla_{\rvz} \Phi(\rvz; \rvx^{(j)}) = \left[\frac{1}{\sigma_{\rvy}^2} A_t^\top (\mathcal{G}(\rvz)-\rvy) + \frac{1}{\rho^2}(\rvz-\rvx^{(j)})\right]$. Then $\pi(\rvz;\rvx^{(j)})$ is an invariant measure of the finite-particle system in Eq.~\eqref{eq:likelihood-dynamics-correction} as well as its large particle limit in Eq.~\eqref{eq:approx-dynamic}. 
\end{lemma}
\begin{proof}
    In the large particle limit, the covariance $C_t$ does not depend on any specific particle but depends on the particle distribution only. Therefore, the Fokker-Plank equation of Eq.~\eqref{eq:approx-dynamic} reads: 
    \begin{align*}
        \frac{\partial}{\partial t} p_t  & = \nabla \cdot \left(p_t C_t \nabla  \Phi(\rvz_t^{(j)}; \rvx^{(j)}) \right) + C_t \nabla^2  p_t \\
        & =  \nabla \cdot \left( p_tC_t \left(\nabla  \Phi(\rvz_t^{(j)}; \rvx^{(j)}) + \nabla\log p_t\right)\right)
    \end{align*}
    where $p_t$ is the probability density at time $t$. We can see that $\pi(\rvz;\rvx^{(j)})$ is an invariant measure by setting both sides to zero. In the finite-particle system, the covariance $C_t=\frac{1}{J} \sum_{j=1}^J(\rvz_t^{(j)}-\bar{\rvz}_t)(\rvz_t^{(j)}-\bar{\rvz}_t)^\top$, which depends on the current state $\rvz_t^{(j)}$. Therefore, the Fokker-Plank equation of the finite-particle dynamics in Eq.~\eqref{eq:likelihood-dynamics-correction} is 
    \begin{align*}
        \frac{\partial}{\partial t} p_t   = & \nabla \cdot \left[p_t \left(C_t \nabla  \Phi(\rvz_t^{(j)}; \rvx^{(j)}) -\frac{n+1}{J}(\rvz_t^{(j)} - \bar{\rvz}_t) \right)\right] + \nabla\cdot\left(\nabla\cdot (p_t C_t)\right) \\
        = & \nabla \cdot \left[p_t \left(C_t \nabla  \Phi(\rvz_t^{(j)}; \rvx^{(j)}) -\frac{n+1}{J}(\rvz_t^{(j)} - \bar{\rvz}_t)\right)\right] + \nabla\cdot \left(C_t \nabla p_t +p_t \nabla\cdot C_t\right) \\
         = & \nabla \cdot \left[p_t C_t\left( \nabla  \Phi(\rvz_t^{(j)}; \rvx^{(j)}) + \nabla \log p_t \right)\right] - \nabla\cdot \left(\frac{n+1}{J}(\rvz_t^{(j)} - \bar{\rvz}_t)\right) + \nabla \cdot \left(p_t \nabla \cdot C_t\right) \\
         = & \nabla \cdot \left[p_t C_t\left( \nabla  \Phi(\rvz_t^{(j)}; \rvx^{(j)}) + \nabla \log p_t \right)\right] - \nabla\cdot \left(\frac{n+1}{J}(\rvz_t^{(j)} - \bar{\rvz}_t)\right)  \\
        & + \nabla \cdot \left(p_t \nabla_{\rvz_t^{(j)}} \cdot \frac{1}{J} \sum_{i=1}^J(\rvz_t^{(i)}-\bar{\rvz}_t)(\rvz_t^{(i)}-\bar{\rvz}_t)^\top \right) \\ 
    = &  \nabla \cdot \left[p_t C_t\left( \nabla  \Phi(\rvz_t^{(j)}; \rvx^{(j)}) + \nabla \log p_t \right)\right] - \nabla\cdot \left(p_t\frac{n+1}{J}(\rvz_t^{(j)} - \bar{\rvz}_t)\right)  \\ 
    & + \nabla \cdot \left( p_t \frac{1}{J}(n+1)(\rvz_t^{(j)} - \bar{\rvz}_t)  \right) \\
    = & \nabla \cdot \left[p_t C_t\left( \nabla  \Phi(\rvz_t^{(j)}; \rvx^{(j)}) + \nabla \log p_t \right)\right]
    \end{align*}
    where the divergence of a matrix is defined in Eq.~\eqref{eq:matrix-divergence}, and we use the following properties: 
    \begin{align*}
        \nabla_{\rvz_t^{(j)}} \cdot \left(\rvz_t^{(j)} \rvz_t^{(j)^\top} \right) & = (n+1)\rvz_t^{(j)}, \\
       \nabla_{\rvz_t^{(j)}} \cdot \left(\rvz_t^{(j)} \rvz_t^{(i)^\top} \right) & = \rvz_t^{(i)}, \\
       \nabla_{\rvz_t^{(j)}} \cdot \left(\rvz_t^{(i)} \rvz_t^{(j)^\top} \right) & = n\rvz_t^{(i)}, \\
       \nabla_{\rvz_t^{(j)}} \cdot \left(\bar{\rvz}_t\bar{\rvz}_t^\top \right) & = \frac{n+1}{J}\bar{\rvz}_t,
    \end{align*}
    where $i\neq j$. By taking both sides to zero, we have that  $\pi(\rvz;\rvx^{(j)})$ is an invariant measure of Eq.~\eqref{eq:likelihood-dynamics-correction}. 
\end{proof}
\begin{remark}
    This proof is largely adapted from~\citet{nusken2019note, garbuno2020affine} which applies to more general scenarios. We tailor and simplify the proof for our use case for the ease of understanding. 
\end{remark}

\section{Practical implementation}
\label{sec:implementation-details}

In this section, we detail the practical implementation of each component of the proposed method, \texttt{Blade}. We also present ablation studies for each design choice to elucidate their individual effects. All the ablation studies are conducted on a small subset of the Navier-Stokes inverse problem's test set. 

\begin{algorithm}
\centering
\caption{Ensemble-based likelihood sampling step}\label{alg:likelihood-step}
\begin{algorithmic}
\STATE {\bfseries Input:} forward model $\mathcal{G}$, observation $\rvy$, effective observation noise $\Tilde{\sigma}_{y}$, coupling strength $\rho$, number of discretization steps $N$, step size scale $\gamma$, initial ensemble $\mathbf{X}=\{\rvx^{(j)}\}_{j=1}^J$, mode (\texttt{main} or \texttt{diag}).
\STATE $\mathbf{Z}_0 \gets \mathbf{X}$
\FOR{$i \in \{0,\dots, N-1\}$}
\STATE $\epsilon_i \sim \mathcal{N}(0, \mI)$
\STATE $\df_1^{(j)} \gets -\frac{1}{\Tilde{\sigma}_y^2}\frac{1}{J} \sum_{k=1}^J \langle \mathcal{G}(\rvz_i^{(k)}) - \bar{\mathcal{G}}, \mathcal{G}(\rvz_i^{(j)}) - \rvy \rangle (\rvz_i^{(k)} - \bar{\rvz}_i),\ j=1,\dots,J$
\IF{mode is \texttt{main}}
\STATE $\df_2^{(j)} \gets -\frac{1}{\rho^2} C_i (\rvx^{(j)} - \rvz_i^{(j)}) + \frac{n+1}{J}(\rvz_i^{(j)}-\bar{\rvz}_i),\ j=1,\dots,J$
\STATE $\sqrt{C_i} \coloneq \frac{1}{\sqrt{J}}\left(\rvz_i^{(1)}-\bar{\rvz}_i, \dots, \rvz_i^{(J)}-\bar{\rvz}_i\right)$
\ELSE
\STATE $\df_2^{(j)} \gets -\frac{1}{\rho^2} C_i (\rvx^{(j)} - \rvz_i^{(j)}),\ j=1,\dots,J$
\STATE $\left(\sqrt{C_i}\right)_k \gets \mathrm{std}\left((\rvz_i)_k\right),\ k=1,\dots,n$
\ENDIF
\STATE $\eta \gets \gamma / \norm{\df_1 + \df_2}_2^2$
\STATE $\rvz_{i + 1}^{(j)} \gets  \rvz_i^{(j)} + (\df_1^{(j)} + \df_2^{(j)}) \eta + \sqrt{2 C_i \eta} \,\epsilon_i,\ j=1,\dots,J$
\ENDFOR
\STATE \textbf{Return} $\mathbf{Z}_N$
\end{algorithmic}
\end{algorithm}

\begin{algorithm}
\centering
\caption{Ensemble prior sampling step}\label{alg:prior-step}
\begin{algorithmic}
\STATE {\bfseries Input:} diffusion model $s_{\theta}$, coupling strength $\rho$, number of discretization steps $N$, initial ensemble $\mathbf{Z}=\{\rvz^{(j)}\}_{j=1}^J$, discretization time steps $t_{i \in \{0, \cdots, N\}}$.
\STATE $\mathbf{X}_{0} \gets \mathbf{Z}$
\STATE $i^* \gets \min\{i \geq 0 \mid \sigma(t_i) \leq \rho\}$
\FOR{$i \in \{i^*,\dots, N - 1\}$}
\IF{\textit{SDE}}
\STATE $\lambda \gets 2$
\ELSE
\STATE $\lambda \gets 1$
\ENDIF
\STATE $\df_i \gets - \lambda t_i s_{\theta}(\mathbf{X}_{i}, \sigma(t_i))$
\STATE $\mathbf{X}_{i + 1} \gets \mathbf{X}_i + (t_{i + 1} - t_i) \df_i$
\IF{$i \neq N - 1$ \textbf{ and } \textit{SDE}}
\STATE $\epsilon_i \sim \mathcal{N}(0, \mI)$
\STATE $\mathbf{X}_{i+1} \gets \mathbf{X}_{i+1} + \sqrt{2 t_i (t_i - t_{i+1})} \,\epsilon_i$
\ENDIF
\ENDFOR
\STATE \textbf{Return} $\mathbf{X}_N$
\end{algorithmic}
\end{algorithm}

\subsection{Likelihood step}

\begin{figure}
    \centering
    \includegraphics[width=0.99\linewidth]{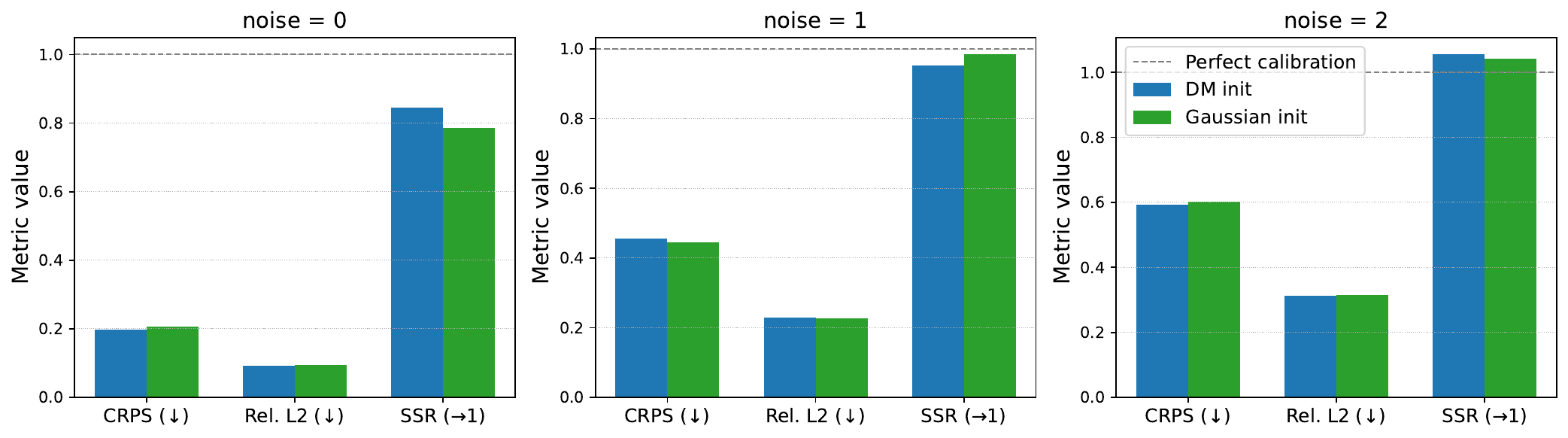}
    \caption{Effect of initialization. At each observation noise level, we compare the performance of \texttt{Blade} initialized from diffusion prior and Gaussian. }
    \label{fig:ablation-init}
\end{figure}

\paragraph{Initialization}
As Theorem~\ref{thm:convergence} indicates, the initialization of \texttt{Blade} is quite flexible, provided the initial distribution maintains a finite KL divergence from the target distribution. Our empirical evaluation considered two initialization strategies: Gaussian and diffusion prior (DM) initialization.
For the Navier-Stokes inverse problem, as shown in Fig.~\ref{fig:ablation-init}, the empirical performance difference between two initializations is not substantial. 
In contrast, for image restoration tasks, we observed that Gaussian initialization gets better results. This is because the target distribution being more closely approximated by a Gaussian due to the large initial coupling strength. Conversely, DM initialization tends to produce a natural image distribution, which is known to often exhibit an unbounded KL divergence from the Gaussian~\citep{arjovsky2017principledmethodstraininggenerative}. Therefore, we use Gaussian initialization for the image restoration tasks. 

\paragraph{Discretization}
We discretize the SDE in Eq.~\eqref{eq:likelihood-dynamics-correction} using the standard Euler method with an adaptive step size defined as
\begin{equation*}
   \mathrm{Step~size} = \gamma \cdot \frac{1}{\| \mathrm{drift} \|_2^2},
\end{equation*}
where $\gamma$ is the hyperparameter that controls the scale of the step size, $\mathrm{drift}$ is the drift term of the SDE in Eq.~\eqref{eq:likelihood-dynamics-correction}. This adaptive step size is effective across all our experiments.  Further design of adaptive step sizes could potentially reduce discretization error with fewer steps. Incorporating the ideas from modern deep learning optimizers for this purpose would be an interesting direction for future work.

\begin{figure}
    \centering
    \includegraphics[width=0.99\linewidth]{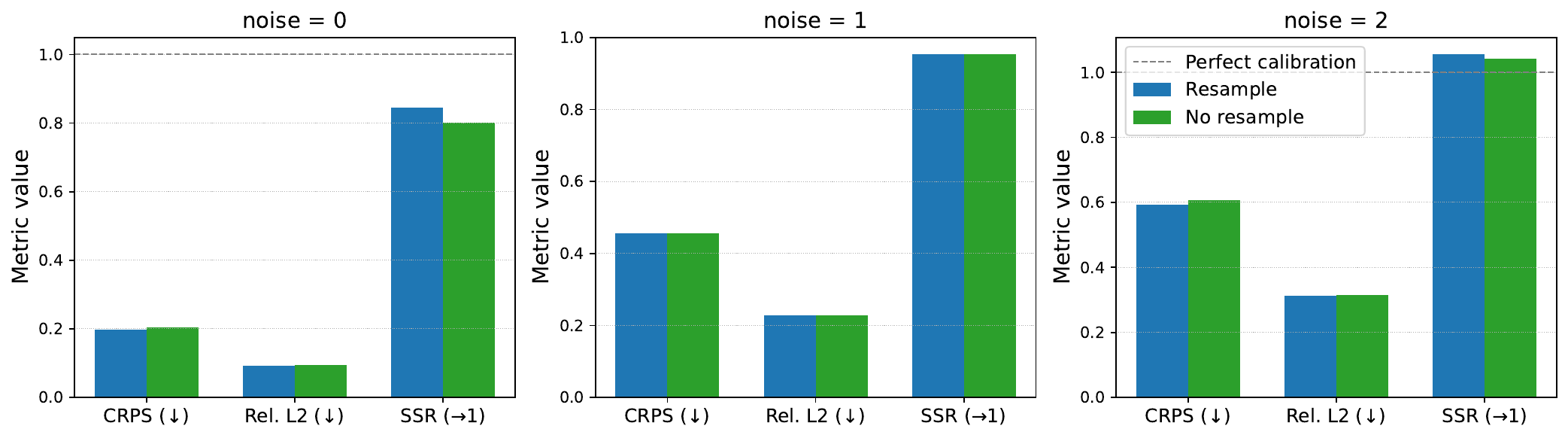}
    \caption{Effect of resampling strategy. At each observation noise level, we compare the performance of \texttt{Blade} with and without the resampling strategy. }
    \label{fig:ablation-resample}
\end{figure}

\paragraph{Resample} \label{sec:resample}
During the likelihood step, we employ a resampling strategy to ensure that the particles are at the correct noise level $\rho$. Resampling is a commonly used method that has been shown to help improve the performance of algorithms such as DAPS \cite{zhang2024improving}, DiffPIR \cite{zhu2023denoising}, and ReSample \cite{song2023solving}. Specifically, we define the following resampling strategy:

\[
\rvz^{(j)}_{\text{resample}} = \rvz^{(j)} + \rho^\prime \epsilon,
\]

where $\epsilon \sim \mathcal{N}(0, \mI)$, $\rho^\prime= \text{max}(0, \rho - \frac{\text{Tr}(C_t)}{n})$, and $n$ is the dimension of the variable $\rvz$. Intuitively, we approximate the current noise level in $\rvz$ and add a corresponding amount of noise to bring $\rvz^{(j)}_{\text{resample}}$ to noise level $\rho$. A key distinction from the prior work is that our $\rho^\prime$ is estimated from the ensemble while the existing methods need to tune it as part of hyperparameters. 
We present the ablation on the effect of resampling strategy in Fig.~\ref{fig:ablation-resample}. As shown, the results with and without resampling are almost the same. In our main experiments, we apply resampling strategy since it introduces minimal additional computation cost and yields slightly better results. 

\begin{figure}
    \centering
    \includegraphics[width=0.75\linewidth]{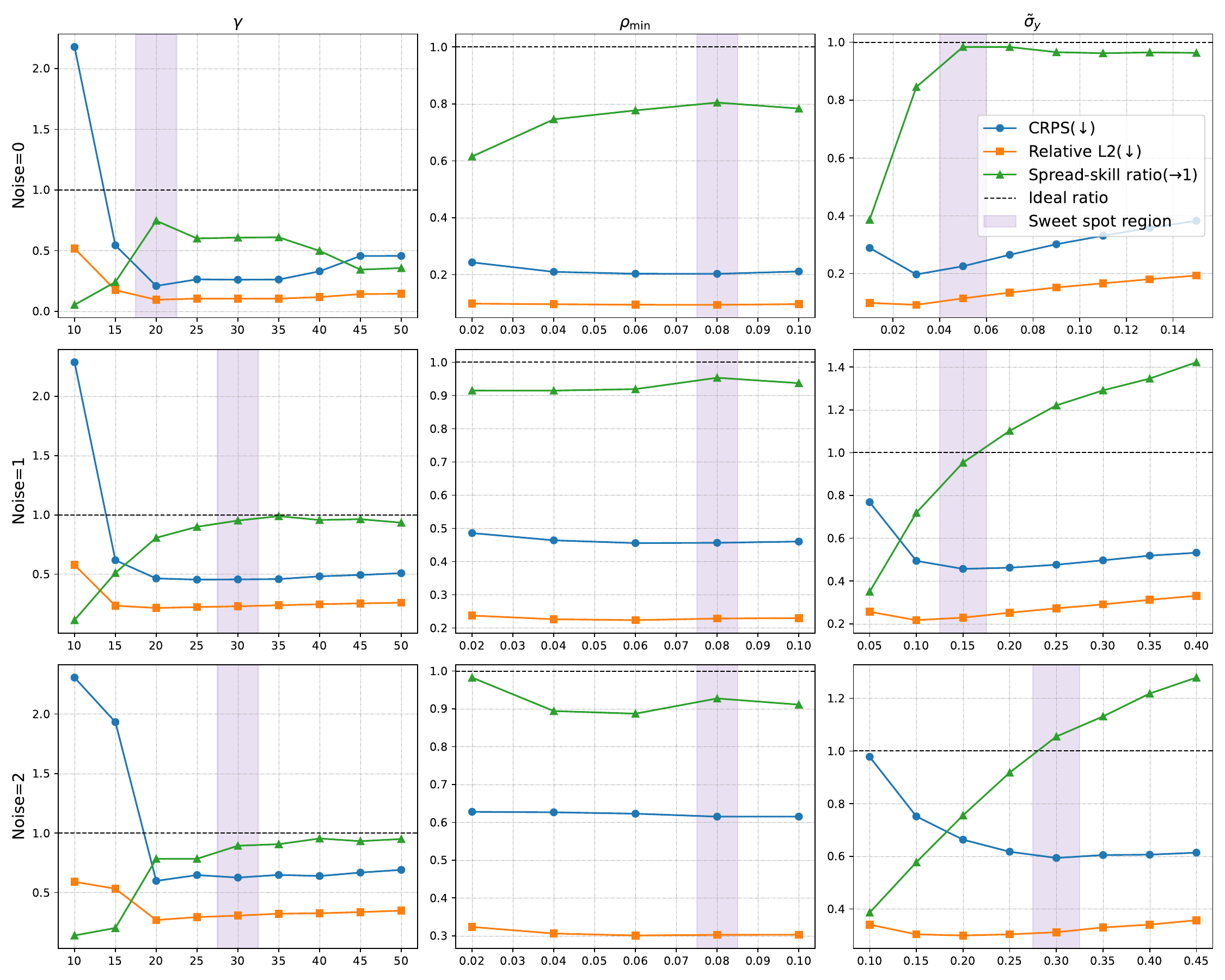}
    \caption{Effect of different hyperparameters of the \texttt{Blade} (main) across different measurement noise levels. }
    \label{fig:ablation-linear-d2}
\end{figure}

\paragraph{Effective observation noise} In practice, we observe that weighting the likelihood with a smaller $\sigma_y$ yields better performance. We denote this adjusted value as the effective observation noise, $\Tilde{\sigma}_y$. Using an effective noise smaller than $\sigma_y$ potentially compensates for the smoothing effect introduced by statistical linearization. In practice, we treat $\Tilde{\sigma}_y$ as a hyperparameter and tune it so that spread-skill ratio is close to 1. The ablation study present in Fig.~\ref{fig:ablation-linear-d2} demonstrates the effect of this hyperparameter. As anticipated, larger $\Tilde{\sigma}_y$ results in an ensemble prediction with greater uncertainty.

\subsection{Prior step}
The prior step is implemented as a denoising diffusion process, with its pseudocode detailed in Algorithm~\ref{alg:prior-step}. We set $\sigma(t)=t$ for simplicity and employ the Euler ODE sampler for faster sampling. 
We discretize the denoising diffusion process with the standard Euler method. Following~\citet{karras2022elucidating}, we use the following step size: 
\begin{equation*}
    t_i = \left(t_{\mathrm{max}}^{1/7} + \frac{i}{N-1}(t_{\mathrm{min}}^{1/7}- t_{\mathrm{max}}^{1/7}) \right)^7, i=0,\dots, N-1.
\end{equation*}

\subsection{Annealing schedule} 
\begin{figure}
    \centering
    \includegraphics[width=0.4\linewidth]{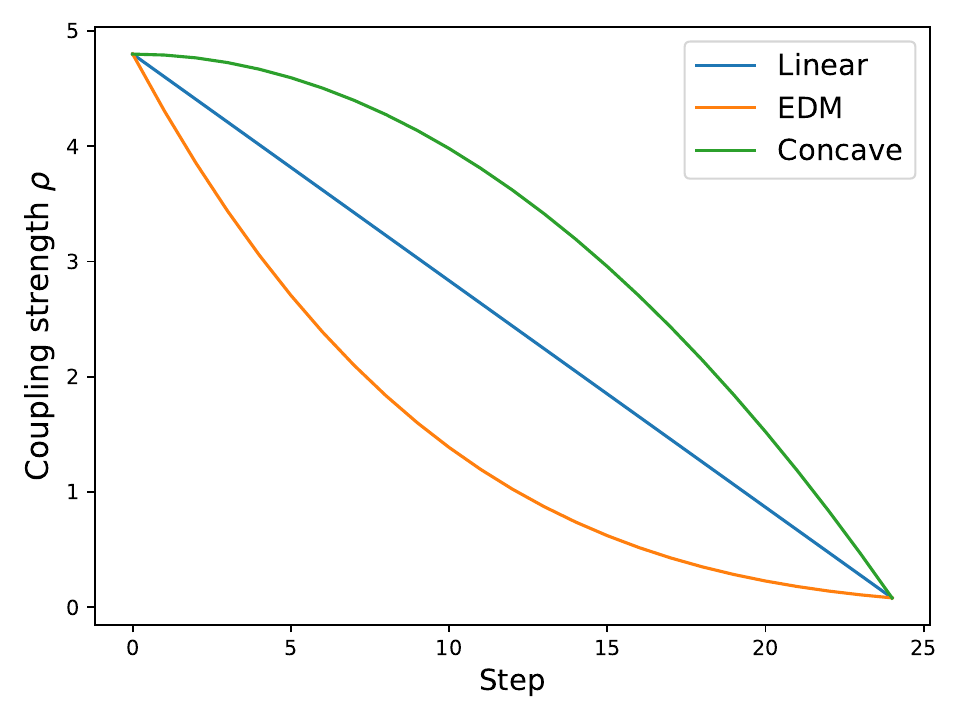}
    \caption{Illustration of the three annealing schedules. Each curve visualizes how the coupling strength $\rho_k$ evolves over iterations.}
    \label{fig:annealing-schedules-visual}
\end{figure}

\begin{figure}
    \centering
    \includegraphics[width=0.75\linewidth]{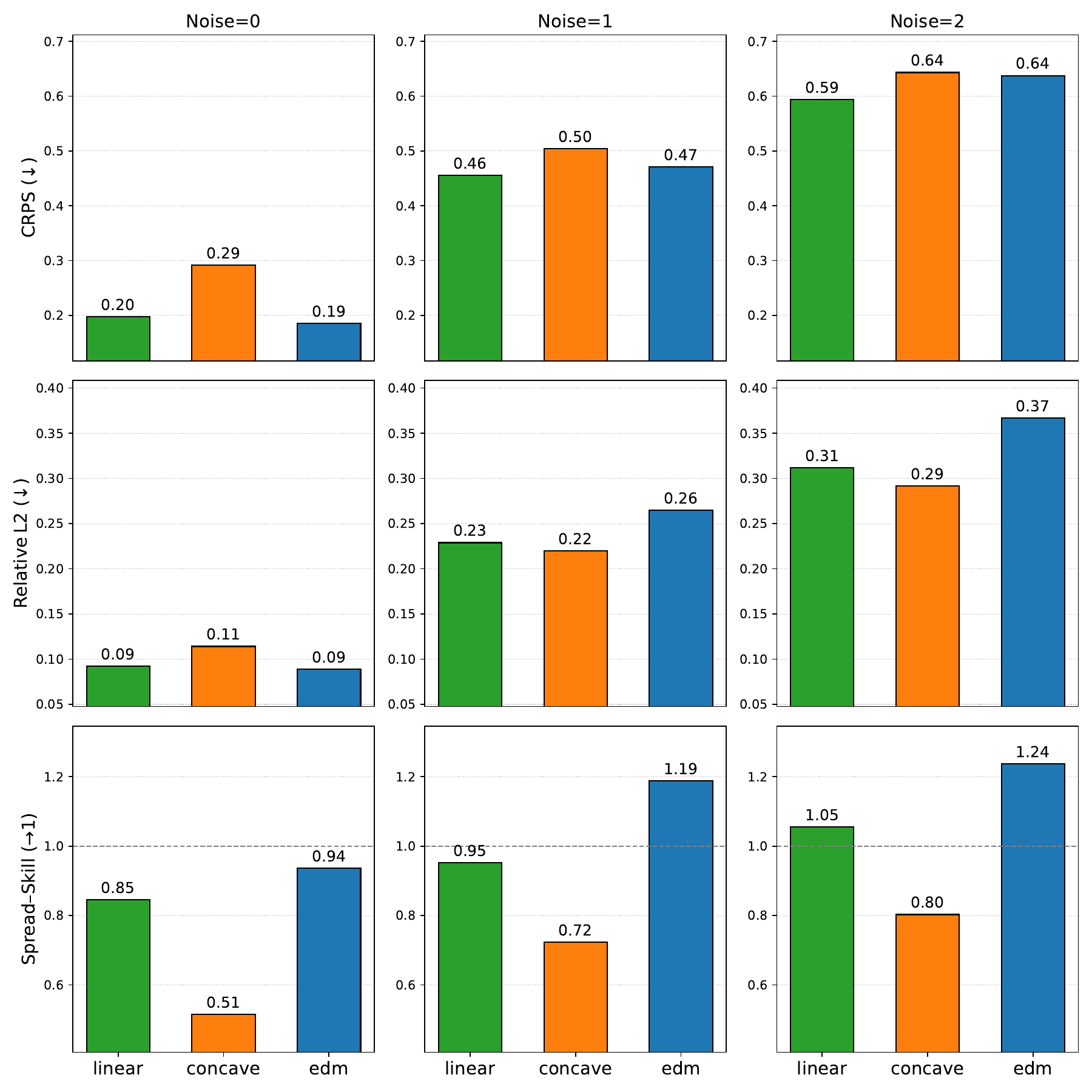}
    \caption{Ablation study on the effect of annealing schedules. }
    \label{fig:ablation-linear-schedule}
\end{figure}

In our experiments, we explored three different types of annealing schedules for the coupling strength $\rho$: linear, EDM, and concave. Given the number of iterations $K$, the maximum value $\rho_\mathrm{max}$, and minimum value $\rho_{\mathrm{min}}$, the linear decay schedule reduces $\rho_k$ as
\begin{equation*}
    \rho_k = \rho_{\mathrm{max}} + \frac{k}{K-1}(\rho_{\mathrm{min}} - \rho_{\mathrm{max}}), k=0,\dots, K-1. 
\end{equation*}
Inspired by the discretization scheme proposed by \citet{karras2022elucidating}, we consider the following schedule, which we refer to as EDM schedule: 
\begin{equation*}
    \rho_k = \left(\rho_{\mathrm{max}}^{1/4} + \frac{k}{K-1}(\rho_{\mathrm{min}}^{1/4}- \rho_{\mathrm{max}}^{1/4}) \right)^4, k=0,\dots, K-1.
\end{equation*}
The quadratic concave schedule is designed to decrease slowly at first and accelerate later, defined as a concave transformation of normalized line: 
\begin{equation*}
    \rho_k = \rho_{\mathrm{min}} + (\rho_{\mathrm{max}} - \rho_{\mathrm{min}})\cdot \left(1 - \frac{k^2}{(K-1)^2}\right), k=0,\dots, K-1. 
\end{equation*}
Fig.~\ref{fig:annealing-schedules-visual} illustrates the behavior of the three annealing schedules described above. The EDM schedule exhibits a steep initial drop, the linear schedule decays uniformly, and the concave schedule maintains a higher coupling strength initially and then decreases rapidly. We study the difference between these schedules for \texttt{Blade (main)} in Fig.~\ref{fig:ablation-linear-schedule}. All three produce reasonable performance, but simple linear schedule offers the most consistent balance of accuracy and calibration across noise levels. Our default is therefore linear.

\begin{figure}
    \centering
    \includegraphics[width=0.75\linewidth]{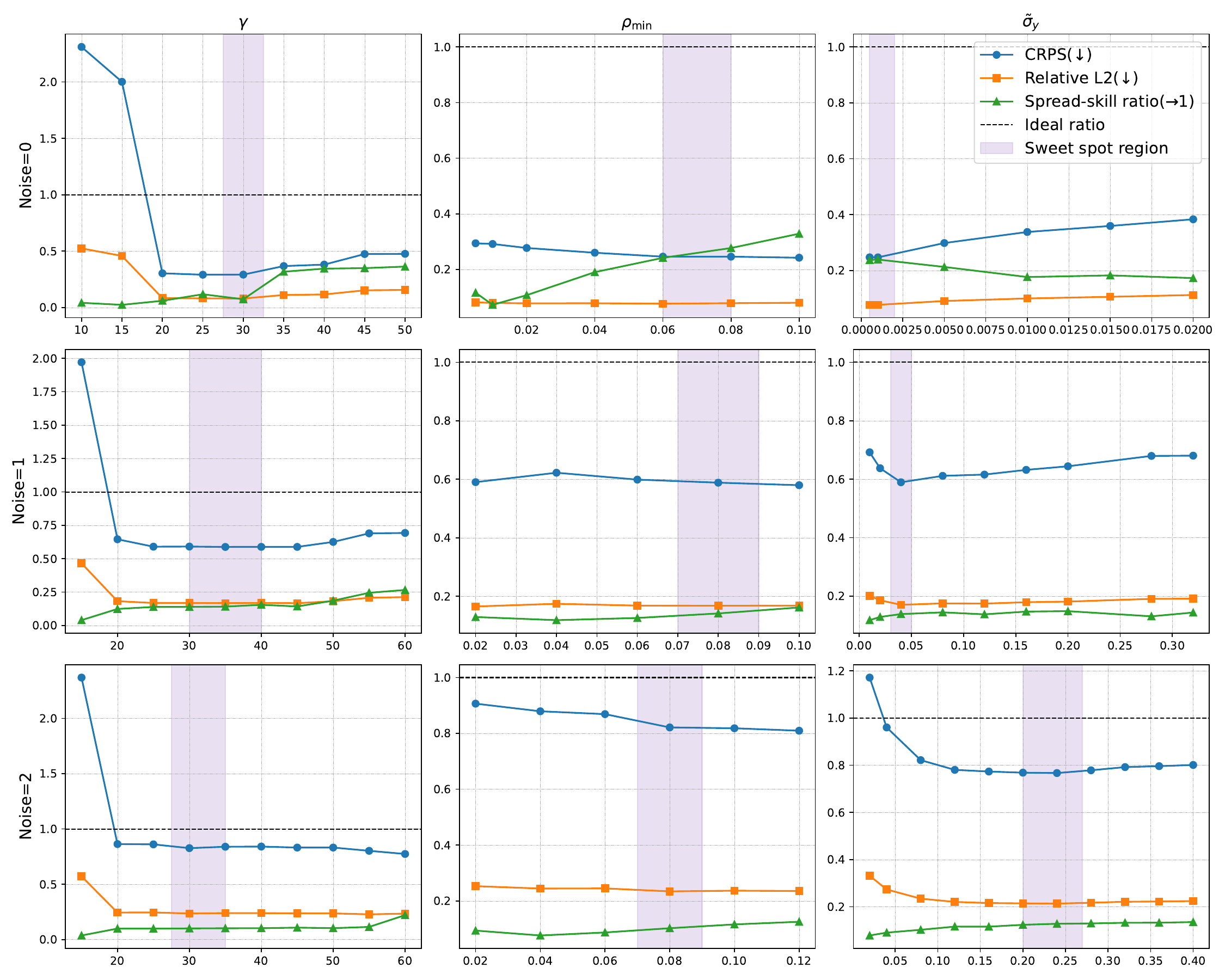}
    \caption{Effect of different hyperparameters of the \texttt{Blade} (diag) across different measurement noise levels. }
    \label{fig:ablation-diag-d2}
\end{figure}

\begin{table}
    \centering
    \caption{Hyperparameter choices of Blade for the main experiments in Table~\ref{tab:navier_stokes}. }
    \label{tab:algo-hyperparameters}
    \resizebox{0.8\linewidth}{!}{
    \begin{tabular}{lcccccccccc}
    \toprule
        Inverse problem & Mode & $\gamma$ & $\Tilde{\sigma}_{y}$ & $\rho_{\mathrm{max}}$ & $\rho_{\mathrm{min}}$  & $\rho$ schedule & $K$ & $N_{\mathrm{likelihood}}$  & $J$ & Init \\ \midrule
         $\sigma_{\mathrm{noise}}=0$ & \texttt{main} &  $20$ & $0.03$ & $4.8$ & $0.08$ & \texttt{linear} & $25$ & $50$ & 512 & \texttt{DM}  \\
             & \texttt{diag} &  30 & 0.001 & $4.8$ & $0.06$ & \texttt{concave} & $25$ & $50$  & 512 & \texttt{DM}  \\ \midrule
         $\sigma_{\mathrm{noise}}=1$ & \texttt{main} &  $30$ & $0.17$ & $4.8$ & $0.08$ & \texttt{linear} & $25$ & $50$  & 512 & \texttt{DM}  \\
           & \texttt{diag}  &  35 & 0.04 & $4.8$ & $0.08$ & \texttt{concave} & $25$ & $50$  & 512 & \texttt{DM}\\ \midrule
          $\sigma_{\mathrm{noise}}=2$ & \texttt{main} &  $30$ & $0.3$ & $4.8$ & $0.08$ & \texttt{linear} & $25$ & $50$  & 512 & \texttt{DM}  \\
          & \texttt{diag} &  30 & 0.25 & $4.8$ & $0.08$ & \texttt{concave} & $25$ & $50$  & 512 & \texttt{DM} \\
         \bottomrule
    \end{tabular}}
\end{table}

\section{Experiment details}

\subsection{Gaussian and Gaussian mixture} \label{sec:exp-linear-gmm-setup}
\subsubsection{Problem setup}
We consider the general linear inverse problem given by
\begin{equation}
    \rvy = H \rvx + \epsilon,
\end{equation}
where $\rvx \in \RR^n, \rvy \in \RR^m, H\in \RR^{m\times n}, \epsilon \sim \normal(0, \Sigma_{\epsilon})$. Given the measurement $\rvy$, we aim to sample from the posterior distribution $p(\rvx|\rvy)$. 
We consider and analyze the case where the prior distribution of $\rvx$ is a mixture of Gaussians given by 
\begin{equation}
    p(\rvx) = \sum_{i=1}^K \gamma_i \normal(m_i,\Sigma_i), \sum_{i=1}^K \gamma_i = 1,  
\end{equation}
where the mean $m_i \in \RR^n$ and the covariance matrix $\Sigma_i\in \RR^{n\times n}$. When $K=1$, the prior degenerates to a Gaussian. 

\paragraph{Linear-Gaussian} In this setting, $K=1$. For all experiments, we randomly generate $m_1$ and choose $\Sigma_i = 25\mI$. We also randomly generate the linear operator $H$. We set $m=1$ and vary $n$.

\paragraph{Linear Gaussian mixture} \label{gmm}
We consider and analyze the case where the prior distribution of $\rvx$ is a mixture of Gaussians given by 
\begin{equation}
    p(\rvx) = \sum_{i=1}^K \gamma_i \normal(m_i,\Sigma_i), \sum_{i=1}^K \gamma_i = 1,  
\end{equation}
where the mean $m_i \in \RR^n$ and the covariance matrix $\Sigma_i\in \RR^{n\times n}$. 

In our experiments, we set the prior to be a mixture of four Gaussians where the variance of each Gaussian is $2\mI$ and the means are $(16i, 16j)$ for $(i,j)\in \{0,1\}^2$. We set $m=1, n=2, \sigma^2_y = 1.5$. The linear forward model $H$ and observed data $\rvy$ are both randomly generated from Gaussian. 

\paragraph{Ground truth posterior} By linearity, the distribution of $\rvy$ is also a Gaussian mixture given by 
\begin{equation}
    p(\rvy) =  \sum_{i=1}^K \gamma_i \normal(Hm_i, H\Sigma_iH^\top + \Sigma_{\epsilon}), \sum_{i=1}^K \gamma_i = 1. 
\end{equation}

Using Bayes theorem, the posterior distribution is given by 
\begin{equation}
    p(\rvx|\rvy) = \frac{p(\rvy|\rvx) p(\rvx)}{p(\rvy)}.
\end{equation}
The likelihood $p(\rvy|\rvx)$ reads 
\begin{equation}
    p(\rvy|\rvx) = \normal(\rvy; H\rvx, \Sigma_{\epsilon}). 
\end{equation}
Therefore, 
\begin{equation}
    p(\rvx|\rvy) = \frac{\sum_{i=1}^K \gamma_i \normal(\rvx; m_i,\Sigma_i)\normal(\rvy; H\rvx, \Sigma_{\epsilon})}{\sum_{i=1}^K \gamma_i \normal(\rvy; m_i, H\Sigma_iH^\top + \Sigma_{\epsilon})}, 
\end{equation}
which can be written as the exponential of a quadratic in $\rvx$. Therefore, the posterior distribution is also a mixture of Gaussians, 
\begin{equation}\label{eq:gmm-posterior}
    p(\rvx|\rvy) = \sum_{i=1}^K \omega_i \normal(\rvx; \hat{m}_i, C_i), 
\end{equation}
where the posterior mean $\hat{m}_i$ and covariance $C_i$ are given by
\begin{align}
    \hat{m}_i & = \left(H^\top \Sigma_\epsilon^{-1}H + \Sigma_i^{-1}\right)^{-1} \left( H^\top\Sigma_\epsilon^{-1}\rvy + \Sigma_i^{-1} m_i\right), \\
    C_i & = \left(H^\top \Sigma_\epsilon^{-1}H + \Sigma_i^{-1}\right)^{-1},
\end{align}
and the weight of each mode is given by 
\begin{equation}
    \omega_j = \frac{\gamma_j \normal(\rvy; Hm_j, H\Sigma_jH^\top + \Sigma_{\epsilon})}{\sum_{i=1}^K \gamma_i \normal(\rvy; Hm_i, H\Sigma_iH^\top + \Sigma_{\epsilon})}, j=1,\dots, K.
\end{equation}

\subsubsection{Evaluation metrics}

\paragraph{KL divergence}

To measure the KL divergence between the distribution of generated samples and the ground truth posterior distribution, we first compute the empirical mean and covariance of the samples. The KL divergence between the $d$-dimensional generated sample distribution $\mathcal{N}(\mu, \Sigma)$ and Gaussian posterior $\mathcal{N}(\mu^*, \Sigma^*)$ is given by

\[
D_{\mathrm{KL}}\left( \mathcal{N}(\mu, \Sigma) \,\|\, \mathcal{N}(\mu^*, \Sigma^*) \right) 
= \frac{1}{2} \left[ 
\log\left( \frac{|\Sigma^*|}{|\Sigma|} \right) 
- d 
+ \mathrm{tr}\left( (\Sigma^*)^{-1} \Sigma \right) 
+ (\mu^* - \mu)^\top (\Sigma^*)^{-1} (\mu^* - \mu)
\right].
\]

The KL divergence helps quantify the error in both mean and covariance (spread) of the generated samples.

\paragraph{W2 distance}

We compute the Sliced Wasserstein Distance between the generated samples and samples from the ground truth posterior:

\[
\text{SWD}_p(P, Q) \approx \left( \frac{1}{L} \sum_{\ell=1}^L W_p^p\left( \langle P, \theta_\ell \rangle, \langle Q, \theta_\ell \rangle \right) \right)^{1/p},
\]

where $P$ is a set of generated samples, $Q$ is the set of samples from the ground truth posterior, $\langle P, \theta_\ell \rangle \text{ and } \langle Q, \theta_\ell \rangle$ denote the empirical 1D distributions formed by projecting the samples onto direction $\theta_\ell$, and $W_p$ is the 1D Wasserstein metric of order $p$. We use the Python Optimal Transport library to compute this metric. 

\subsection{Navier-Stokes equation} \label{sec:navier-stokes-setup}
\subsubsection{Problem setup}

Following the experimental setup in InverseBench~\citep{zhenginversebench}, we consider the 2-d Navier-Stokes equation for a viscous, incompressible fluid in vorticity form on a torus,
\begin{align}
\label{eq:ns}
\begin{split}
\partial_t \rvw(\rvx,t) + \rvu(\rvx,t) \cdot \nabla \rvw(\rvx,t) &= \nu \Delta \rvw(\rvx,t) + f(\rvx), \qquad \rvx \in (0,2\pi)^2, t \in (0,T]  \\
\nabla \cdot \rvu(\rvx,t) &= 0, \qquad \qquad \qquad \qquad \quad\, \rvx \in (0,2\pi)^2, t \in [0,T]  \\
\rvw(\rvx,0) &= \rvw_0(\rvx), \qquad \qquad \qquad \quad \rvx \in (0,2\pi)^2 
\end{split}
\end{align}
where $\rvu$ is the velocity field, 
$\rvw = \nabla \times \rvu$ is the vorticity, $\rvw_0 $ is the initial vorticity,  $ \nu\in \RR_+$ is the viscosity coefficient, and $f$ is the forcing function. 
The solution operator $\mathcal{F}$ is defined as the operator mapping the vorticity from the initial vorticity to the vorticity at time $T$.
$\mathcal{F}: \rvw_0 \rightarrow \rvw_T$. Numerically, it is realized as a pseudo-spectral solver~\citep{he2007stability}. This Navier-Stokes equation is a standard benchmark problem widely used in the literature~\citep{iglesias2013ensemble,li2020fourier, takamoto2022pdebench}.
The forward model in our inverse problem is given by 
\begin{equation}
    \rvy = PL(\mathcal{F}(\rvw_0)) + \epsilon,
\end{equation}
where $L$ is the discretization operator and $P$ is the sampling operator.


\subsubsection{Evaluation metrics}\label{sec:metrics}
We adopt the following three standard metrics to evaluate the results from different perspectives. 


\paragraph{Relative L2 error}
Suppose $\rvx^*$ is the ground truth function and $\rvx$ is the predicted function. The relative L2 error measures the error $\rvx - \rvx^*$ relative to the norm of the ground truth: 
\begin{equation*}
    \mathrm{Rel~L2~error} = \frac{\|\rvx - \rvx^*\|_2}{\|\rvx^*\|_2}.
\end{equation*}

\paragraph{Continuous Ranked Probability Score (CRPS)} The CRPS~\citep{gneiting2007strictly} is a standard probabilistic metric to assess the quality of the entire predicted distribution for inverse problems, which is defined as 
\begin{equation*}
    \mathrm{CRPS} = \EE | \rvx - \rvx^* | - \frac{1}{2}\EE |\rvx - \rvx^\prime|, 
\end{equation*}
where $\rvx, \rvx^\prime$ are independent random predictions and $\rvx^*$ is the single observed ground truth. Intuitively, it measures the distance between a predicted distribution and the single observed ground truth $\rvx^*$ that actually occurred. It is minimized when the ensemble prediction is drawn from the same distribution as the ground truth, i.e., $ \rvx^{(j)} \sim p(\rvx^*)$ for all $j$. We consider the multi-dimensional version of CRPS defined in~\citet{rasp2024weatherbench}. For an ensemble prediction $\{\rvx^{(j)}\}_{j=1}^J$ where $\rvx^{(j)}\in\RR^n$, the CRPS for the single ground truth $\rvx^*$ is given by 
\begin{equation}
     \mathrm{CRPS} = \frac{1}{n}\sum_{i=1}^{n} \left(\frac{1}{J} \sum_{j=1}^{J} |\rvx^{(j)}(i) - \rvx^*(i)| - \frac{1}{2J(J-1)} \sum_{j=1}^{J}\sum_{k=1}^{J}|\rvx^{(j)}(i) - \rvx^{(k)}(i) | \right),
\end{equation}
which can be implemented in $O(nJ\log J)$ complexity using the equivalent form introduced in~\citet{zamo2018estimation}. In our experiments, we report the CRPS averaged over all test cases.

\paragraph{Spread-skill ratio (SSR)}
The spread-skill ratio (SSR) is a simple yet powerful diagnostic of how well an ensemble prediction's stated uncertainty (spread) matches its actual error (skill)~\citep{fortin2014should}. Intuitively, if the ensemble distribution truly captures the variability of the ground truth, then ensemble members should be statistically indistinguishable from observed outcomes. Formally, let  $\{\rvx^*_{i}\}_{i=1}^N$ denote a set of observed ground truths. Suppose, for each observed ground truth $\rvx^*_i$, we have an ensemble prediction $\{\rvx_{i,j}\}_{j=1}^J$. Let $\bar{\rvx}_i=\frac{1}{J}\sum_j \rvx_{i,j}$. The unbiased estimator of SSR can be written as 
\begin{equation}
    \mathrm{SSR} = \sqrt{\frac{\mathrm{spread}^2}{\mathrm{skill^2}}}, 
\end{equation}
where 
\begin{align*}
    \mathrm{spread}^2 & = \frac{1}{N} \sum_{i=1}^N  \frac{1}{J-1} \sum_{j=1}^J\|\rvx_{i,j} - \bar{\rvx}_{i}\|_2^2, \\ 
    \mathrm{skill}^2 & = \frac{1}{N}\sum_{i=1}^N\|\frac{1}{J}\sum_j \rvx_{i,j} -\rvx^*_{i} \|^2_2 + \frac{1}{J(J-1)} \mathrm{spread}^2.
\end{align*}
A value of $\mathrm{SSR}=1$ indicates the perfect calibration. Small $\mathrm{SSR}$ means that the ensemble prediction is over-confident while large $\mathrm{SSR}$ indicates that the ensemble prediction is over-cautious. 

\paragraph{Rank histogram}

The rank histogram~\citep{anderson1996method,hamill1997reliability,talagrand1999evaluation} assesses ensemble calibration by comparing the truth to the empirical distribution formed by the ensemble. For each grid point in each test case, the ensemble members are sorted, and the rank of the true value within this ordering ($0$ means below all, $J$ means above all, or an intermediate integer) is recorded. Pooling these ranks over all points and cases yields a histogram with $J + 1$ bins. A flat histogram indicates a statistically consistent ensemble: the truth behaves like an additional random draw from the predictive distribution. A U-shape signals under-dispersion (ensemble spread too narrow), an inverted U indicates over-dispersion, and tilted shapes reveal bias.

\subsubsection{Baseline implementation}\label{sec:baseline-details}
For methods that require training on paired data, specifically the end-to-end U-Net and conditional diffusion model (CDM), we first generate a collection of observation-solution pairs by simulating observations from the prior training dataset available in InverseBench~\citep{zhenginversebench}. To evaluate their in-distribution performance, we retrain the U-Net and CDM for each noise level, which takes around 7-10 hours on a single GH200.
\paragraph{U-Net} 
The end-to-end U-Net baseline is adapted from the U-Net used in our diffusion model by removing the time conditioning branch. The observation is upsampled to the same resolution before being fed into the U-Net. However, it is important to note that observations are not always spatially aligned with the unknown signal in a general setting. Consequently, end-to-end neural networks typically require additional design considerations for different types of observations.

\paragraph{CDM-CA} 
CDM-CA is adapted from the U-Net architecture of the prior diffusion model. This involved replacing the self-attention module with cross-attention and incorporating a CNN-based observation encoder, following the conditioning mechanism used in~\citet{rombach2022high}. 
\paragraph{CDM-Cat} CDM-Cat is adapted from the U-Net of the prior diffusion model. We upsample the observation to the same resolution as the solution and directly concatenate it with the input over the channel dimension. This conditioning mechanism only works for observations that are spatially aligned to the input.  

To find the right hyperparameters for the training-free methods, we perform two rounds of Bayesian optimization over their hyperparameters on the validation set provided in InverseBench~\citep{zhenginversebench}.

\paragraph{Existing baselines in InverseBench} We simply take the existing implementations from InverseBench~\citep{zhenginversebench} for baseline methods that already exist in the codebase including DPG~\citep{tang2024solving}, SCG~\citep{huang2024symbolic}, EnKG~\citep{zheng2024ensemble}, and EKI~\citep{iglesias2013ensemble}. 

\paragraph{EKS + DM} The original EKS~\citep{garbuno2020interacting} method only considers the Gaussian prior. The most straightforward way to incorporate the diffusion prior is to initialize the ensemble members with random samples from diffusion model. We fix the ensemble size to 1024 and optimize the other hyperparameters using Bayesian optimization.

\paragraph{EKS w/ DP} While the original update rules of EKS is derived for Gaussian prior only, we can approximate the gradient of the prior regularization term in EKS by evaluating the diffusion model $s_{\theta}(\rvx_t; \sigma(t))$ with $t$ close to zero. A learned score is least reliable in this small-noise regime, so this approximation can bias the prior gradient---unlike \texttt{Blade}, which uses the diffusion model only as a sampler. We fix the ensemble size to 1024 and optimize the other hyperparameters using Bayesian optimization. 

\paragraph{Localized EKS w/ DP} Localization is a common technique to improve the multi-modal sampling of ensemble methods by assigning a localized mean and covariance for each particle. We implement the localized EKS w/ DP following the formulation in ~\citep{reich2021fokker,wagner2022ensemble}. Again, we fix the ensemble size to 1024 and optimize the other hyperparameters using Bayesian optimization. 

\paragraph{FKD} Feynman-Kac diffusion steering (FKD)~\citep{zhao2025conditional,singhal2025general} is a method based on interacting particle system, which can be viewed as a generalization of SCG~\citep{huang2024symbolic}. We fix the ensemble size to 1024 and optimize the other hyperparameters using Bayesian optimization. 

\subsection{Image restoration}\label{sec:image-appendix}

\subsubsection{Problem setup} \label{sec:image-setup}


We evaluated our method on image restoration tasks with the FFHQ256 dataset. Our evaluation set consisted of the first ten images, indexed 00000 to 00009, in the validation set. The pre-trained model is taken from \citet{chung2023diffusion} (FFHQ256) and converted into an EDM checkpoint with their Variance-Preserving (VP) preconditioning \citep{karras2022elucidating}. In general, we follow the experiment setup of ~\citet{chung2023diffusion}. All problem settings use a measurement noise of $\sigma_y = 0.05$. 
We address that PnP-DM results on phase retrieval are significantly worse than those originally reported. Note that we used a different forward model configuration, larger measurement noise, and full-color images, which differs from the original PnP-DM setup. Furthermore, PnP-DM super-resolution results are lower than originally reported. We note that we compare against the PnP-DM configuration that uses Langevin Monte Carlo during the likelihood step, which differs from their original configuration for linear inverse problems.

\paragraph{Box Inpainting}

The forward model is given by 

\[
\rvy \sim \mathcal{N}(\textit{\textbf{M}} \odot \rvx, \sigma_{\rvy}^2 \mI),
\]

where $\textit{\textbf{M}}$ is a binary masking matrix. We consider the case where the mask is a boxed region, which requires significantly stronger guidance from the prior for generation of plausible image content.

\paragraph{Super-resolution}

The forward model is given by 

\[
\rvy \sim \mathcal{N}(\textit{\textbf{P}}_f \rvx, \sigma_{\rvy}^2 \mI),
\]

where $\textit{\textbf{P}}_f$ is a linear operator that downsamples an image by a factor of $f$ with a block averaging filter. In our experiments, we set $f$ = 4. 

\paragraph{Phase Retrieval}

The forward model is given by 

\[
\rvy \sim \mathcal{N}(\lvert \textit{\textbf{F}} \textit{\textbf{P}} \rvx \rvert, \sigma_{\rvy}^2 \mI),
\]

where $\textit{\textbf{P}}$ is an oversampling matrix and $\textit{\textbf{F}}$ is the Fourier transform. We set the oversampling ratio to $2$. 

\subsubsection{Evaluation metrics}

\paragraph{PSNR}

Peak Signal-to-Noise Ratio (PSNR) measures the ratio between the maximum power of a signal and the maximum power of the noise corrupting it. PSNR is a commonly used metric to assess the quality of image and video reconstruction. The PSNR between a prediction $\rvx$ and ground truth signal $\rvx^*$ is defined as 

\begin{equation}
    \mathrm{PSNR} = 20 \cdot \log_{10} (\mathrm{MAX}_{\rvx}) - 10 \cdot \log_{10} (\mathrm{MSE}(\rvx, \rvx^*)),
\end{equation}

where $\mathrm{MAX}_{\rvx}$ is the maximum possible pixel value (i.e. 255). 

\paragraph{SSIM}

The structural similarity index measure (SSIM) ~\citep{wang2004image} is another metric to compute the similarity between two images. It compares patterns of luminance, contrast, and structure between two images to achieve a metric more aligned with visual perception, given by 

\begin{equation}
    \mathrm{SSIM} = \frac{(2 \mu_\rvx \mu_{\rvx^*} + C_1)(2 \sigma_{\rvx \rvx^*} + C_2)}{(\mu_\rvx^2 + \mu_{\rvx^*}^2 + C_1)(\sigma_{\rvx}^2 + \sigma_{\rvx^*}^2 + C_2)},
\end{equation}

where $\mu_\rvx, \mu_{\rvx^*}$ are the mean luminances, $\sigma_\rvx, \sigma_{\rvx^*}$ the the variances between images $\rvx$ and $\rvx^*$, respectively. The terms $C_1$ and $C_2$ are small constants to stabilize the computation and $\sigma_{\rvx \rvx^*}$ the covariance between the two images.

\paragraph{LPIPS}

The Learned Perceptual Image Patch Similarity (LPIPS) ~\citep{zhang2018unreasonable} compares two images by measuring their distance in the feature space of a pre-trained neural network. LPIPS is computed as a weighted sum of the Euclidean distances between the network activations of an image $\rvx$ and an image $\rvx^*$.

\subsection{Black hole imaging}\label{sec:bh-appendix}

\subsubsection{Problem setup}
We use the black hole imaging benchmark from InverseBench~\citep{zhenginversebench}, which recovers a black hole image from the sparse, noisy measurements of the Event Horizon Telescope. To cancel per-telescope calibration errors, the forward model is built from closure quantities, which are nonlinear and make the problem non-convex. The diffusion prior is trained on $64\times64$ GRMHD-simulated black-hole images; we refer to InverseBench for the full measurement model.

\subsubsection{Evaluation metrics}
We report PSNR and Blur PSNR (PSNR after blurring to the telescope's resolution), and measure data consistency with the closure-phase and closure-amplitude $\chi^2$ statistics ($\chi^2_{cp}$, $\chi^2_{camp}$; ideal value $1$).



\section{Additional experiments}
\paragraph{Linear Gaussian problems} We provide additional quantitative results on linear Gaussian problems. We report the SWD and $\KL$ for each method across different problem dimensions and observation noise levels in Table~\ref{tab:LG}. As shown, \texttt{Blade}, EKS and MCGDiff~\citep{cardoso2024monte} both achieve similar and strong performance, surpassing the other tested baselines. This aligns with theoretical expectations, given that these methods are proven to be asymptotically accurate in linear Gaussian problems. Consequently, the results empirically confirm that \texttt{Blade} can achieve accurate posterior sampling when applied to linear problems.

\label{sec:additional-results}

\begin{figure}
    \centering
    \includegraphics[width=0.99\linewidth]{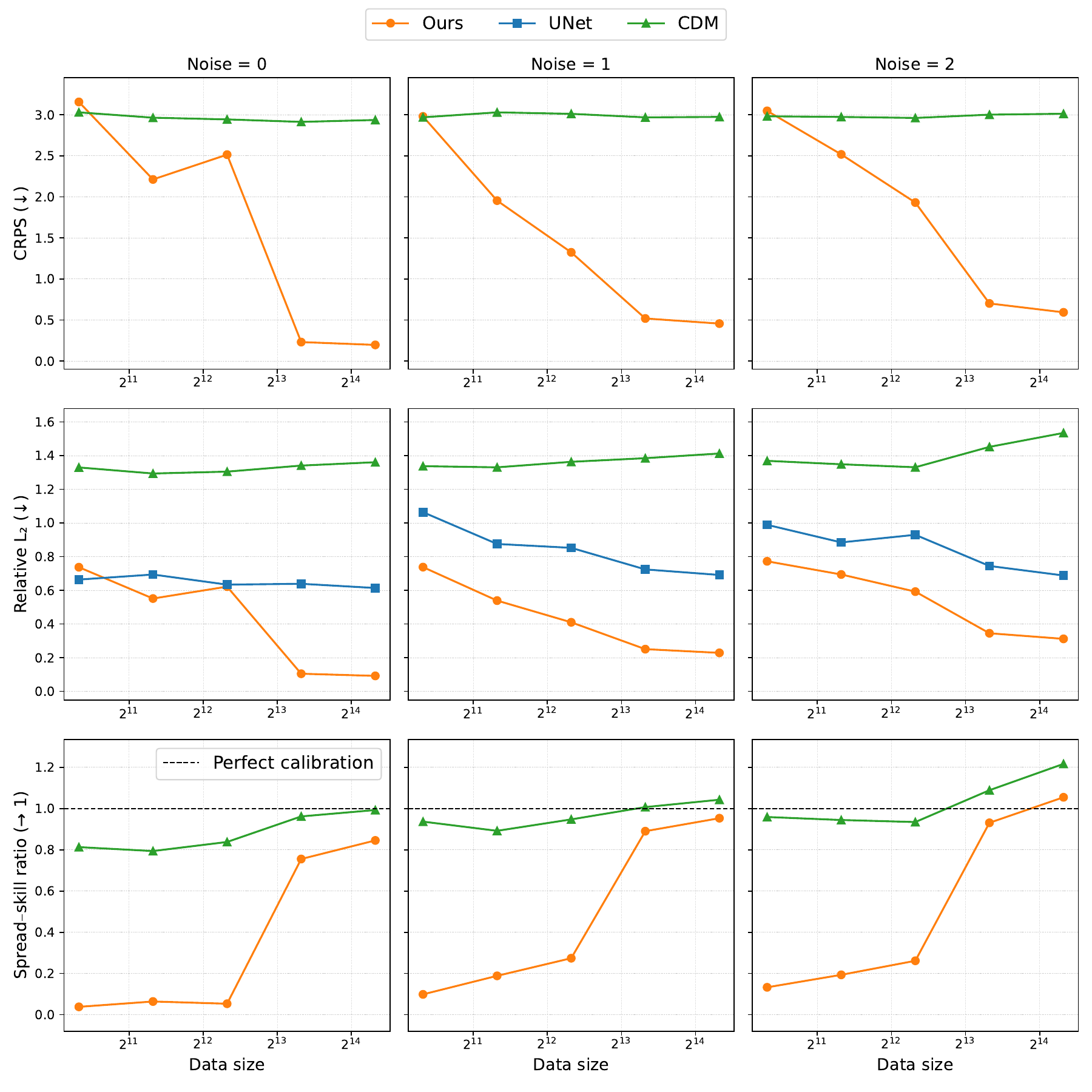}
    \caption{Scaling of different methods' performance with training data size across three measurement noise levels. Each row corresponds to one evaluation metric. U-Net produces deterministic prediction so the probabilistic metrics like CRPS and SSR do not apply. Note that our method uses the same diffusion prior across different noise levels while the training-based methods train separate networks for each noise level. }
    \label{fig:data-scaling}
\end{figure}

\begin{table}[ht]
    \centering
    \caption{Experimental results on linear Gaussian problems for Sliced Wasserstein Distance (SWD) and KL Divergence ($\KL$) across different \( \sigma_y \) values and methods for various data dimension $n$.}
    \label{tab:LG}
    \resizebox{0.8\linewidth}{!}{
    \small
    \begin{tabular}{cccccccccc}
        \toprule
        \multirow{2}{*}{\( n \)} & \multirow{2}{*}{Method} & \multicolumn{8}{c}{\( \sigma_y \)} \\
        \cmidrule(lr){3-10}
         & & \multicolumn{2}{c}{0.5} & \multicolumn{2}{c}{1.5} & \multicolumn{2}{c}{2.5} & \multicolumn{2}{c}{3.5} \\
        \cmidrule(lr){3-4} \cmidrule(lr){5-6} \cmidrule(lr){7-8} \cmidrule(lr){9-10}
         & & SWD & $\KL$ & SWD & $\KL$ & SWD & $\KL$ & SWD & $\KL$ \\
        \midrule
        \multirow{4}{*}{2} & DPG  & 4.199 & 11.83 & 3.955 & 122.11 & 3.969 & 350.76 & 4.646 & 1219.96 \\
         & SCG  & 2.826 & 85k & 2.704 & $\geq$ 100k & 3.072 & $\geq$ 100k & 3.814 & $\geq$ 100k \\
         & EnKG & {1.832} & $\geq$ 100k & {1.752} & $\geq$ 100k & {1.972} & $\geq$ 100k & 3.020 & $\geq$ 100k \\
         & EKS & \underline{1.651} & \textbf{0.374} & 2.072 & \underline{0.493} & 2.061 & \underline{0.502} & 2.204 & \underline{0.549} \\
         & MCGdiff & \textbf{1.423} & 1.002 & \textbf{1.497} & 1.238 & \textbf{1.511} & 0.899 & \underline{1.760} & 0.985 \\
         & Blade (Ours) & 1.915 & \underline{0.556} & \underline{1.725} & \textbf{0.348} & \underline{1.763} & \textbf{0.423} & \textbf{1.678} & \textbf{0.381} \\
        \midrule
        \multirow{4}{*}{80} & DPG  & 6.786 & 69.66 & 7.005 & 67.32 & 6.905 & \underline{66.11} & 7.289 & 67.13 \\
         & SCG  & 6.022 & 1708. & 6.059 & 15916. & 6.033 & 37121. & 6.013 & 91466. \\
         & EnKG & 4.997 & $\geq$ 100k & 4.938 & $\geq$ 100k & 5.180 & $\geq$ 100k & 5.068 & $\geq$ 100k \\
         & EKS & \textbf{2.444} & \textbf{27.72} & \textbf{2.357} & \textbf{25.91} & \textbf{2.366} & \textbf{25.69} & \textbf{2.371} & \textbf{25.60} \\
         & MCGdiff & 33.03 & 177.3 & 32.90 & 177.2 & 32.87 & 177.2 & 32.93 & 177.5 \\
         & Blade (Ours) & \underline{4.367} & \underline{64.13} & \underline{4.492} & \underline{65.99} & \underline{4.284} & 67.65 & \underline{4.579} & \underline{61.06} \\
        \midrule
        \multirow{4}{*}{400} & DPG  & 6.111 & $\geq$ 100k & 6.149 & $\geq$ 100k & 6.259 & $\geq$ 100k & 6.181 & $\geq$ 100k \\
         & SCG  & 6.199 & $\geq$ 100k & 6.172 & $\geq$ 100k & 6.182 & $\geq$ 100k & 6.276 & $\geq$ 100k \\
         & EnKG & 8.636 & $\geq$ 100k & 7.432 & $\geq$ 100k & 11.513 & $\geq$ 100k & 11.825 & $\geq$ 100k \\
         & EKS & \textbf{1.047} & \textbf{1817.} & \textbf{1.092} & \underline{1777.} & \textbf{1.114} & \underline{2156.} & \textbf{1.130} & \underline{2134.} \\
         & MCGdiff & 33.09 & $\geq$ 100k & 33.08 & $\geq$ 100k & 32.98 & $\geq$ 100k & 33.06 & $\geq$ 100k \\
         & Blade (Ours) & \underline{4.414} & \underline{2478.} & \underline{4.527} & \textbf{1747.} & \underline{4.074} & \textbf{1794.} & \underline{4.340} & \textbf{1805.} \\
        \bottomrule
    \end{tabular}}
\end{table}


\begin{figure}
    \centering
    \includegraphics[width=0.5\linewidth]{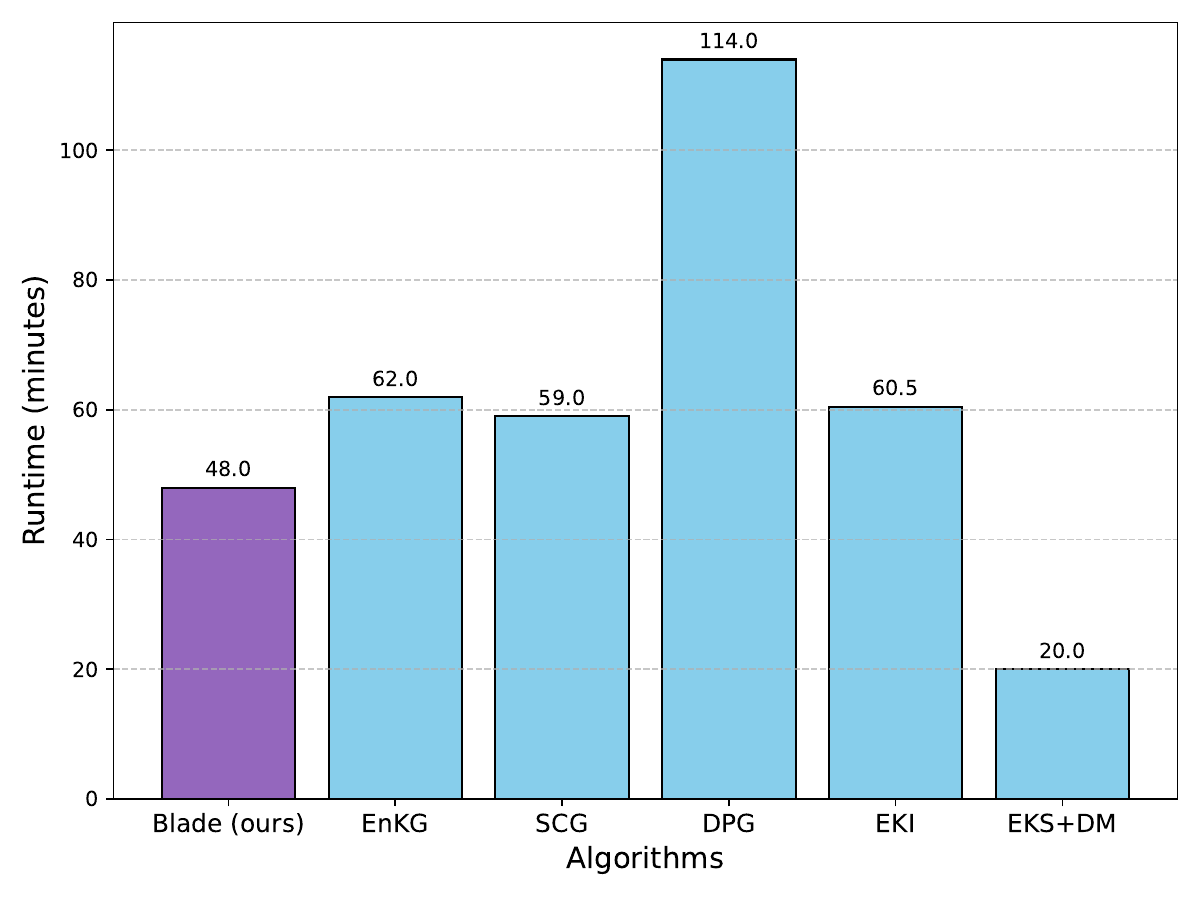}
    \caption{Runtime comparison of different algorithms on Navier-Stokes inverse problem, measured on a single GH200. }
    \label{fig:runtime-comparison}
\end{figure}

\end{document}